\PassOptionsToPackage{table,dvipsnames,svgnames,x11names}{xcolor}

\documentclass[sigconf, nonacm]{acmart}

\usepackage{tikz} 
\usepackage{multirow}
\usepackage{enumitem}
\usepackage{booktabs}
\usepackage{balance}
\usepackage{algorithm}
\usepackage[noend]{algpseudocode}
\usepackage{multicol}
\usepackage{hyperref}

\usepackage{array}
\usepackage{tcolorbox}
\tcbuselibrary{skins} 
\usetikzlibrary{positioning,fit,backgrounds,calc,shadows,shapes.multipart}
\definecolor{gold}{RGB}{255, 215, 0}
\fontsize{8.8}{8}\selectfont
\newcommand{\SAGA}{\mathsf{SagaLLM}}

\newcommand\vldbdoi{XX.XX/XXX.XX}
\newcommand\vldbpages{XXX-XXX}
\newcommand\vldbvolume{14}
\newcommand\vldbissue{1}
\newcommand\vldbyear{2025}
\newcommand\vldbauthors{\authors}
\newcommand\vldbtitle{\shorttitle} 
\newcommand\vldbavailabilityurl{URL_TO_YOUR_ARTIFACTS}
\newcommand\vldbpagestyle{plain} 

\begin{document}
\title{SagaLLM: Context Management, Validation, and Transaction Guarantees for Multi-Agent LLM Planning}

\author{Edward Y. Chang}
\affiliation{%
  \institution{Stanford University}
}
\email{echang@cs.stanford.edu}

\author{Longling Geng}
\affiliation{%
  \institution{Stanford University}
}
\email{gll2027@stanford.edu}

\begin{abstract}
This paper introduces SagaLLM, a structured multi-agent architecture designed to address four foundational limitations of current LLM-based planning systems: unreliable self-validation, context loss, lack of transactional safeguards, and insufficient inter-agent coordination. While recent frameworks leverage LLMs for task decomposition and multi-agent communication, they often fail to ensure consistency, rollback, or constraint satisfaction across distributed workflows.
SagaLLM bridges this gap by integrating the Saga transactional pattern with persistent memory, automated compensation, and independent validation agents. It leverages LLMs’ generative reasoning to automate key tasks traditionally requiring hand-coded coordination logic, including state tracking, dependency analysis, log schema generation, and recovery orchestration. Although SagaLLM relaxes strict ACID guarantees, it ensures workflow-wide consistency and recovery through modular checkpointing and compensable execution.
Empirical evaluations across planning domains demonstrate that standalone LLMs frequently violate interdependent constraints or fail to recover from disruptions. In contrast, SagaLLM achieves significant improvements in consistency, validation accuracy, and adaptive coordination under uncertainty—establishing a robust foundation for real-world, scalable LLM-based multi-agent systems.
\end{abstract}

\maketitle

\pagestyle{\vldbpagestyle}
\begingroup\small\noindent\raggedright\textbf{PVLDB Reference Format:}\\
\vldbauthors. \vldbtitle. PVLDB, \vldbvolume(\vldbissue): \vldbpages, \vldbyear.\\
\href{https://doi.org/\vldbdoi}{doi:\vldbdoi}
\endgroup
\begingroup
\renewcommand\thefootnote{}\footnote{\noindent
This work is licensed under the Creative Commons BY-NC-ND 4.0 International License. Visit \url{https://creativecommons.org/licenses/by-nc-nd/4.0/} to view a copy of this license. For any use beyond those covered by this license, obtain permission by emailing \href{mailto:info@vldb.org}{info@vldb.org}. Copyright is held by the owner/author(s). Publication rights licensed to the VLDB Endowment. \\
\raggedright Proceedings of the VLDB Endowment, Vol. \vldbvolume, No. \vldbissue\ %
ISSN 2150-8097. \\
\href{https://doi.org/\vldbdoi}{doi:\vldbdoi} \\
}\addtocounter{footnote}{-1}\endgroup


\ifdefempty{\vldbavailabilityurl}{}{
\vspace{.3cm}
\begingroup\small\noindent\raggedright\textbf{PVLDB Artifact Availability:}\\
The source code, data, and/or other artifacts have been made available at \url{\vldbavailabilityurl}.
\endgroup
}

\section{Introduction}
\label{sec:intro}

Multi-Agent Systems (MAS) have long been a cornerstone of distributed computing and database systems \cite{lesser2004distributed, jennings1995commitments, DuninKeplicz2010TeamworkIM}. Over the past few decades, their development has followed two primary trajectories. In the database community, MAS traditionally integrated the foundational transaction processing principles, particularly ACID properties \cite{Gray81, weikum2001transactional}, to ensure consistency and reliability for complex multi-step operations. For long-lived, distributed, or loosely-coupled tasks, MAS also adopted more flexible transactional models (e.g., Saga \cite{HectorSaga1987}) to maintain robustness while relaxing strict atomicity or isolation constraints.

Parallel to these database-oriented approaches, distributed systems research emphasized coordination protocols and flexible collaboration mechanisms \cite{wooldridge2009introduction, durfee1999distributed}, enabling scalable multi-agent interactions without the overhead of strict locking or heavyweight transactional guarantees. These complementary development paths have resulted in frameworks optimized for different priorities: \emph{transactional integrity} versus \emph{adaptive coordination}, highlighting the fundamental trade-off between strong consistency and flexible execution in practical systems.

Recent advances in Large Language Models (LLMs) \cite{vaswani2017attention, LLMSurvey101145ACMTrans, zhao2025surveylargelanguagemodels} have revitalized MAS as a paradigm for sophisticated reasoning and multi-agent collaboration \cite{du2023improvingfactualityreasoninglanguage, SocraSynthChangCSCI2023, EVINCEChang2024}. Frameworks such as AutoGen \cite{wu2024autogen}, LangGraph \cite{langgraph2024}, and CAMEL \cite{li2023camel} demonstrate how LLM-based agents can decompose tasks, interact between modalities, and coordinate to solve complex problems. However, this resurgence often neglects the foundational transaction guarantees essential to reliable multi-agent workflows, particularly in domains requiring robust state management.

Unlike traditional MAS, LLM-based systems often lack mechanisms for maintaining strong consistency, failure recovery, and rollback handling, leading to inconsistent states, partial failures, and unreliable execution in real-world applications. These limitations stem from several fundamental challenges: LLMs struggle with \emph{internal validation} due to inherent limitations highlighted by Gödel's incompleteness theorems \cite{godel1931english}, making them unreliable for detecting and correcting their own errors. Furthermore, \emph{context loss} in long conversations \cite{modarressi2025nolimalLONGContext, xiao2024attentionsink, hsieh2024lostinthemiddle, liu-etal-2024-lost} can cause LLMs to forget earlier steps, leading to contradictory decisions. When tasks are distributed across multiple agents, these problems are compounded, as no built-in supervisory mechanism exists to reconcile state changes or validate constraint satisfaction across agents.

For example, in a travel booking scenario, an LLM-based MAS can independently issue flight and hotel reservations without ensuring their coordinated success. If the flight is later canceled, the system may not recognize the inconsistency, leaving the hotel reservation active. Such scenarios illustrate the critical need for transactional frameworks that preserve the intelligence and adaptability of LLM-based MAS while ensuring consistency and reliability in long-running, interdependent workflows.

To address these limitations, we propose $\SAGA$, a multi-agent transactional system that extends the Saga pattern, a transactional model originally developed to manage complex, long-lived transactions by decomposing them into smaller, independently validated, committed, and compensable units. By integrating transactional logic, compensatory rollback mechanisms, and \emph{persistent memory} into LLM-based MAS, $\SAGA$ ensures that each individual operation within a workflow is reliably validated and committed, with clearly defined compensating transactions that restore system-wide consistency in case of failure.

Crucially, $\SAGA$ leverages the reasoning and coding capabilities of LLMs to automate core aspects of transaction orchestration that previously required scenario-specific manual programming. $\SAGA$ enables LLMs to:
\begin{enumerate}[leftmargin=1.7em, topsep=.1em, parsep=.1em, label=\arabic*.]
\item Identify the persistent states to track,
\item Validate constraints and inter-agent dependencies,
\item Design logging schemas to capture workflow transitions,
\item Develop compensatory logic for failure recovery, and
\item Implement communication protocols among agents to coordinate these behaviors.
\end{enumerate}
Traditionally, each of these components required custom implementation by system developers, tailored to individual applications. In contrast, $\SAGA$ employs LLM as intelligent agents that automatically infer, generate, and coordinate these mechanisms, greatly improving scalability and reducing development overhead.

This hybrid approach---integrating transactional processing with adaptive multi-agent intelligence---makes $\SAGA$ particularly effective for real-world applications that are complex and demand reliability and safety in e.g., healthcare management, supply chain management, and emergency response. These capabilities collectively enable $\SAGA$ to overcome foundational limitations in LLM-based MAS, such as unreliable state coordination, lack of rollback support, and limited context retention, thereby supporting robust, scalable reasoning across distributed agents. We summarize our key contributions below:

\begin{enumerate}[leftmargin=1.7em, topsep=.1em, parsep=.1em, label=\arabic*.]
\item \textbf{Transactional Consistency via Persistent Memory and Compensation}: $\SAGA$ introduces transactional safeguards and persistent-memory-based compensatory mechanisms to LLM-based MAS, ensuring reliable consistency and coherent state recovery across multi-agent workflows.
\item \textbf{Robust Constraint and Dependency Validation}: $\SAGA$ incorporates temporal-spatial context tracking and external verification mechanisms to validate inter-agent dependencies and prevent inconsistencies, addressing the fundamental limits of LLM self-verification.
\item \textbf{LLM-Orchestrated Intelligence}: $\SAGA$ automates key components of multi-agent planning---state tracking, constraint checking, log schema design, compensation logic, and coordination protocols---through the generative reasoning and coding capabilities of LLMs.
\end{enumerate}

The remainder of this paper covers: related work (Section~\ref{sec:related}), problem definition (Section~\ref{sec:requirements}), architecture (Section~\ref{sec:design}), evaluation (Section~\ref{sec:exp}), and conclusions (Section~\ref{sec:conc}).
\section{Related Work}
\label{sec:related}

We review three strands of related work: (1) the evolution of transactional management, particularly the Saga pattern; (2) cognitive limitations of LLMs motivating the need for transactional integrity, independent validation, and context preservation; and (3) multi-agent LLM frameworks and recent attempts to integrate transactional safeguards.

\subsection{Transaction Management Systems}
\label{sec:transaction-evolution}

Transactional models have evolved significantly since Gray introduced the ACID properties \cite{Gray81}. In distributed settings, strict ACID guarantees became impractical, prompting models such as BASE \cite{pritchett2008base} and long-lived transaction patterns.

The Saga pattern by Garcia-Molina and Salem \cite{HectorSaga1987} decomposes long-lived transactions into smaller, locally atomic sub-transactions with compensating steps for failure recovery. This has influenced modern microservice architectures and workflow engines \cite{richardson2018microservices}.

Systems such as YAWL \cite{vanderAalst2005YAWLYA}, AWS Step Functions \cite{aws-step-functions}, and Azure Logic Apps \cite{azure-logic-apps} embed Saga-style workflows but remain rigid and manually defined, lacking dynamic adaptability. These foundational principles inform $\SAGA$’s adaptive, LLM-driven extensions.

\subsection{LLM Limitations Necessitating SagaLLM's Key Requirements}
\label{sec:llm-limitations}

To realize this complementary relationship between AI and workflow systems, we must first address the specific limitations that prevent current LLMs from functioning effectively in transactional workflows.
Our analysis reveals three fundamental limitations that directly motivate $\SAGA$'s core requirements: transactional integrity, independent validation, and strategic context preservation.

\vspace{-.05in}
\paragraph{\textbf{Self-Validation Gap Necessitating Independent Validation}} LLMs inherently lack robust self-validation mechanisms, a limitation originating from intrinsic boundaries identified by Gödel's incompleteness theorems, demonstrating fundamental constraints on a system's ability to verify its own reasoning \cite{godel1931english, chang2025ALAS}. Recent research confirms that self-refinement techniques \cite{madaan2022,li2023dissectingCOT,jiang2024selfincorrect}, while iterative and beneficial, are unable to surpass inherent capability ceilings to reliably correct deeper logical errors \cite{huang2024large}. In transactional scenarios, these validation gaps manifest as factual inconsistencies, invalid operations, and unreliable plan feasibility assessments \cite{yamin2024failuremodesllmscausal}. Thus, $\SAGA$ incorporates an independent validation framework to mitigate these inherent limitations.

\vspace{-.05in}
\paragraph{\textbf{Statelessness Necessitating Transactional Integrity}} LLMs process each interaction independently, lacking native mechanisms to maintain the state across sequential interactions. This fundamental statelessness necessitates explicit transactional integrity management to maintain coherent operation sequences and ensure robust failure recovery. Without systematic transaction management, LLM-based systems risk state inconsistency, operation losses, and incoherent recovery procedures.

\vspace{-.05in}
\paragraph{\textbf{Context Limitations and Strategic Preservation}}
LLMs rely on self-attention mechanisms that prioritize recent tokens, leading to significant degradation in context retention over long sequences. Empirical studies reveal sharp drops in recall beyond token limits \cite{xiao2024attentionsink, modarressi2025nolimalLONGContext}, especially for mid-context information \cite{liu-etal-2024-lost, hsieh2024lostinthemiddle}. Chain-of-thought heuristics \cite{cemri2025multiagentllmsystemsfail} further exacerbate this by lacking mechanisms to manage or pass context reliably across steps. These limitations hinder multi-step reasoning \cite{WeiCOT3600270, stechly2024COTLimits}, as earlier outputs are frequently lost. $\SAGA$ addresses this by explicitly preserving vulnerable context elements—goals, justifications, and dependencies—through structured memory and persistent tracking.

Collectively, these limitations provide strong motivation to address all three key requirements within $\SAGA$. Comprehensive transaction management, independent validation, and strategic context preservation are essential for reliably deploying LLM-based multi-agent systems (MAS) in critical real-world applications.

\subsection{Multi-Agent LLM Frameworks and Transaction Limitations}
\label{sec:mas-limitations}

Frameworks such as AutoGen \cite{wu2024autogen}, LangGraph \cite{langgraph2024}, and CAMEL \cite{li2023camel} advance multi-agent LLM coordination but fall short in addressing $\SAGA$'s three core requirements: transactional integrity, independent validation, and context preservation.

\paragraph{\textbf{Missing Transaction Semantics}} LangGraph and AutoGen enable structured workflows and agent interactions, but lack built-in atomicity guarantees, compensation logic, or robust failure recovery. AgentScope \cite{gao2024agentscope2024} and AFlow \cite{zhang2024aflow} introduce limited rollback mechanisms, but do not generalize across workflows.

\paragraph{\textbf{Validation Gaps}} Most frameworks are based on LLM self-validation, exposing them to reasoning errors and hallucinations. Systems such as PLASMA \cite{PLASMA2024procedural} improve reliability but omit transaction-level rollback. LLM-MCTS \cite{zhao2023LLM-MCTS} and Tree-of-Thought \cite{yao2024tree} emphasize pre-execution reasoning without runtime consistency checks.

\paragraph{\textbf{Limited Context Preservation}} CAMEL preserves dialogue history but lacks mechanisms for tracking state transitions, inter-agent dependencies, or compensatory paths. Broader planning systems \cite{wei2025plangenllmsmodernsurveyllm} do not offer persistent context retention strategies.
\\
\newline
Unlike these systems, $\SAGA$ treats compensation, validation, and context tracking as first-class design goals—ensuring reliable, recoverable, and intelligent coordination across complex multi-agent workflows.

\section{System Requirements}
\label{sec:requirements}

Building on the limitations identified in Section~\ref{sec:related}, this section formally defines $\SAGA$'s requirements. While $\SAGA$ inherits the core transactional semantics of classical Saga patterns \cite{HectorSaga1987}, significant adaptations are required for LLM-based multi-agent systems, as summarized in Table~\ref{tab:saga-comparison}.

To address foundational limitations in LLM-based execution, $\SAGA$ is organized around three tightly interwoven yet conceptually distinct requirements that extend classical transaction processing into the realm of adaptive multi-agent intelligence:

\begin{enumerate}[leftmargin=1.7em, label=\arabic*.]
\item \textbf{Transactional Integrity:}
Ensures that agent operations transition the system through coherent, globally consistent states. This is achieved through structured rollback mechanisms, compensating actions, and invariant preservation across interdependent agents. It also requires reliable tracking of system state to detect and repair inconsistencies triggered by partial execution or disruptions.

\item \textbf{Independent Validation:}
Addresses the known limitations of LLM self-verification by introducing cross-agent and external validation layers. These mechanisms evaluate agent outputs and inter-agent inputs against constraints, schemas, and dependency graphs. Persistent validation histories are maintained to support rollback triggers and guard against hallucinations or invalid commitments.

\item \textbf{Context Management:}
Maintains essential state and dependency information across long-horizon workflows. Rather than relying on ephemeral context windows, $\SAGA$ persistently stores goals, justifications, and compensation plans in structured memory, enabling agents to reason over consistent histories and perform accurate recovery after failures.
\end{enumerate}

This integrated design contrasts with prior systems that handle these aspects in isolation or without formal guarantees. The mutual reinforcement among these three pillars reflects the central challenge of LLM-based workflows: effective orchestration under uncertainty, without sacrificing consistency and correctness.

\begin{table}[t]
\centering
\fontsize{8}{9}\selectfont
\caption{Classical Saga vs $\SAGA$ Framework Comparison}
\vspace{-.1in}
\label{tab:saga-comparison}
\begin{tabular}{>{\raggedright\arraybackslash}p{1.7cm}>{\raggedright\arraybackslash}p{2.2cm}>{\raggedright\arraybackslash}p{3.0cm}}
\toprule
\textbf{Aspect} & \textbf{Classical Saga} & \textbf{$\SAGA$} \\
\midrule
Domain & Database transactions & Multi-agent LLM workflows \\
\midrule
Compensation & Pre-defined rollback procedures & LLM-generated + validated compensation \\
\midrule
Validation & Schema/constraint validation & Independent LLM output validation \\
\midrule
Context & Stateless transactions & Strategic context preservation \\
\midrule
Coordination & Simple sequential execution & Complex multi-agent dependency management \\
\midrule
Intelligence & Rule-based workflows & Adaptive LLM reasoning w/ transaction guarantees \\
\bottomrule
\end{tabular}
\vspace{-.1in}
\end{table}

\subsection{Transactional Integrity Requirements}

$\SAGA$ provides transactional guarantees tailored for multi-agent workflows, extending classical transaction semantics across autonomous agent boundaries. Sequences $O = \{o_1, o_2, ..., o_n\}$ are operations treated as a unit of logical cohesiveness, where each $o_i$ is locally atomic. If any operation fails, $\SAGA$ initiates compensatory actions to restore global consistency.

Applying $O$ to a system state $S$ must yield either a fully committed state $S'$, or trigger a coherent rollback that returns the system to $S$, thereby avoiding partial or inconsistent outcomes. To ensure this, $\SAGA$ enforces the following properties:

\paragraph{\textbf{Consistency Preservation}} $\SAGA$ ensures that all state transitions respect global invariants $I$. If $S \models I$, then any resulting $S' \models I$, even when execution spans multiple agents.

\paragraph{\textbf{Isolation Guarantees}} $\SAGA$ guarantees that concurrently executing agents produce final states equivalent to some serial order, regardless of autonomy or internal decision processes.

\paragraph{\textbf{Durability Assurance}} $\SAGA$ guarantees persistence of committed states by durably recording execution outcomes and metadata necessary for fault recovery and compensatory execution.

To enforce these guarantees, $\SAGA$ maps each $o_i$ to a local transaction $T_i$, paired with a compensating transaction $C_i$. In case of failure at step $T_j$, compensating actions are invoked in reverse:

\begin{equation}
\text{Saga } S = \{T_1, T_2, ..., T_n, C_n, ..., C_2, C_1\}.
\end{equation}

\subsubsection{Transaction State Management}

$\SAGA$ maintains a structured state representation across three orthogonal dimensions to support validation, compensation, and recovery:

\begin{itemize}[leftmargin=1.5em, topsep=.05em, parsep=.05em]
\item \textit{Application State ($S_A$):} Domain-specific entities and system checkpoints
\item \textit{Operation State ($S_O$):} Execution logs, inputs, outputs, and LLM reasoning traces
\item \textit{Dependency State ($S_D$):} Graph-structured constraints and satisfaction criteria
\end{itemize}

\begin{table}[t]
\centering
\caption{Transaction State Management in $\SAGA$}
\fontsize{8.5}{10}\selectfont
\vspace{-.1in}
\label{tab:transaction-state}
\begin{tabular}{>{\raggedright\arraybackslash}p{2.5cm}>{\raggedright\arraybackslash}p{4.2cm}}
\toprule
\textbf{Mechanism} & \textbf{Information Recorded} \\
\hline
\midrule
\multicolumn{2}{l}{\textit{\textbf{Application State ($S_A$)}}} \\
\midrule
Domain Entities & 
\begin{minipage}[t]{\linewidth}
\begin{itemize}[nosep,leftmargin=*]
    \item Application-domain objects
    \item Entity states and status
    \item Checkpoints and snapshots
\end{itemize}
\end{minipage} \\
\midrule
\multicolumn{2}{l}{\textit{\textbf{Operation State ($S_O$)}}} \\
\midrule
Execution Logs & 
\begin{minipage}[t]{\linewidth}
\begin{itemize}[nosep,leftmargin=*]
    \item Operation inputs and outputs
    \item Timestamps and execution status
    \item Completion indicators
\end{itemize}
\end{minipage} \\
Decision Reasoning & 
\begin{minipage}[t]{\linewidth}
\begin{itemize}[nosep,leftmargin=*]
    \item LLM-generated reasoning chains
    \item Justifications and alternatives
\end{itemize}
\end{minipage} \\
Compensation Metadata & 
\begin{minipage}[t]{\linewidth}
\begin{itemize}[nosep,leftmargin=*]
    \item Inverse operations
    \item Preconditions and recovery state
\end{itemize}
\end{minipage} \\
\midrule
\multicolumn{2}{l}{\textit{\textbf{Dependency State ($S_D$)}}} \\
\midrule
Causal Dependencies & 
\begin{minipage}[t]{\linewidth}
\begin{itemize}[nosep,leftmargin=*]
    \item Inter-operation constraints
    \item Data and resource flow mappings
\end{itemize}
\end{minipage} \\
Constraint Satisfaction & 
\begin{minipage}[t]{\linewidth}
\begin{itemize}[nosep,leftmargin=*]
    \item Boolean condition checks
    \item Satisfaction evidence and timestamps
\end{itemize}
\end{minipage} \\
\bottomrule
\end{tabular}
\end{table}

\subsubsection{Dependency Tracking and Compensation Planning}

$\SAGA$ models operation dependencies as a directed graph:
\begin{equation}
D = \{(o_i, o_j, c_{ij}) \mid o_j \text{ depends on } o_i \text{ under condition } c_{ij}\}.
\end{equation}
To express more complex conditions:
\begin{equation}
c_{\{i_1, ..., i_n\}, j} = \mathcal{B}(c_{i_1 j}, ..., c_{i_n j}),
\end{equation}
where $\mathcal{B}$ is a Boolean function over prerequisite conditions.

Upon failure, $\SAGA$ traverses this graph to determine the minimal set of affected operations and executes compensatory actions that restore global consistency without violating preserved invariants.

\subsection{Independent Validation Requirements}

To address the inherent limitations of LLM self-verification, \textsc{SAGA} introduces a two-tier validation architecture governed by a global validation agent. This agent operates independently of task agents and has visibility into the full transaction history, agent communications, and global state.

\paragraph{\textbf{Intra-Agent Output Validation}} The global validation agent inspects the outputs of the individual task agents before those outputs are committed or transmitted. Outputs are checked for:
\begin{itemize}[leftmargin=1.5em, topsep=.05em, parsep=.05em,label=-]
\item Syntactic correctness (format, schema)
\item Semantic coherence and reasoning soundness
\item Factual accuracy against context
\item Constraint adherence and invariants
\item Context preservation and dependency awareness
\end{itemize}
Failures trigger compensations or corrective augmentation.

\paragraph{\textbf{Inter-Agent Input and Dependency Validation}} Inputs and messages between agents are validated before delivery. Checks include:
\begin{itemize}[leftmargin=1.5em, topsep=.05em, parsep=.05em, label=-]
\item Contract conformance
\item Dependency satisfaction
\item Cross-agent consistency
\item Temporal ordering
\item Mutual agreement on shared state
\item Transaction coherence
\end{itemize}
Failed validations block delivery and invoke recovery.

\paragraph{\textbf{Validation Response Protocols}} \textsc{SAGA} defines structured validation outcomes:
\begin{itemize}[leftmargin=1.5em, topsep=.05em, parsep=.05em, label=-]
\item \textit{Rejection:} Discard and compensate
\item \textit{Augmentation:} Enhance with clarifications
\item \textit{Feedback:} Record for future adaptation
\end{itemize}

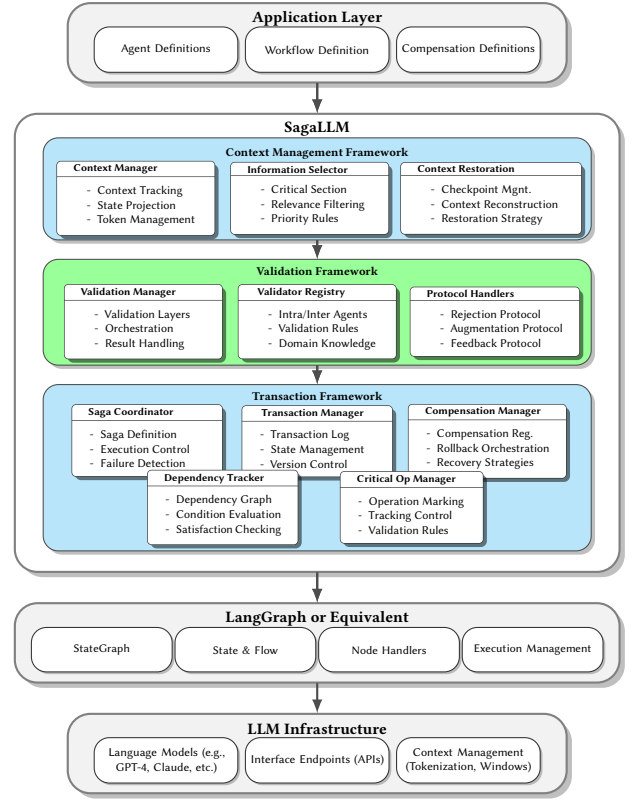
\begin{figure}[t!]
\vspace{-.1in}
\begin{center}
\begin{tikzpicture}[scale=0.5, transform shape,
    font=\sffamily\normalsize,
    layer/.style={
        rectangle,
        rounded corners=3mm,
        draw=black!50,
        thick,
        fill=#1,
        drop shadow={shadow xshift=0.7mm, shadow yshift=-0.7mm},
        text width=15.6cm,
        align=center,
        inner sep=2mm
    },
    comp_group/.style={
        rectangle,
        rounded corners=2mm,
        draw=black!70,
        fill=#1,
        drop shadow={shadow xshift=0.5mm, shadow yshift=-0.5mm},
        text width=14.2cm,
        minimum height=5cm,
        align=center,
        inner sep=1.5mm,
        font = \Large
    },
    component/.style={
        rectangle,
        rounded corners=2mm,
        draw=black!80,
        fill=white,
        drop shadow={shadow xshift=0.3mm, shadow yshift=-0.3mm},
        text width=3.6cm,
        minimum height=1.2cm,
        align=center,
        inner sep=1mm
    },
    feature_list/.style={
    text width=4.5cm,
    align=left,
    font=\small,
    itemize/.style={leftmargin=3em}  
    },
    layer_title/.style={
        font=\normalsize\bfseries
    },
    group_title/.style={
        font=\normalsize\bfseries
    },
    arrow/.style={
        -latex,
        thick,
        draw=black!70
    }
]

\node[layer=gray!10, minimum height=2.1cm, text width=12.8cm] (app_layer) {};
\node[layer_title] at ($(app_layer.north)-(0,0.4cm)$) {\LARGE Application Layer};

\node[component] (agent_def) at ($(app_layer.center)+(-4.0cm,-.18)$) {\normalsize Agent Definitions};
\node[component] (workflow_def) at ($(app_layer.center)+(0.0cm,-.18)$) {\normalsize Workflow Definition};
\node[component] (comp_def) at ($(app_layer.center)+(4.0cm,-.18)$) {\normalsize Compensation Definitions};

\node[layer=white!10, minimum height=12.12cm, below=0.8cm of app_layer] (sagallm) {};
\node[layer_title] at ($(sagallm.north)-(0,0.4cm)$) {\LARGE SagaLLM};

\node[comp_group=cyan!25, minimum height=2.7cm] (context_pres) at ($(sagallm.north)-(0,2cm)$) {};
\node[group_title] at ($(context_pres.north)-(0,0.4cm)$) {\normalsize Context Management Framework};

\node at ($(context_pres.center)+(-4.8cm,-0.25cm)$) {
    \begin{tcolorbox}[
        width=4.2cm,
        title=Context Manager,
        fonttitle=\small\bfseries,
        colbacktitle=white,
        coltitle=black,
        colback=white,
        colframe=black!80,
        arc=1mm,
        boxrule=0.5pt,
        boxsep=1pt,
        toptitle=1pt,
        bottomtitle=1pt,
        enhanced,
        drop shadow={shadow xshift=0.5mm, shadow yshift=-0.5mm}
    ]
    \begin{itemize}[leftmargin=2em, topsep=0pt, label=-, nosep]
        \item Context Tracking
        \item State Projection
        \item Token Management
    \end{itemize}
    \end{tcolorbox}
};
\node at ($(context_pres.center)+(-0.2cm,-0.25cm)$) {
    \begin{tcolorbox}[
        width=4.2cm,
        title=Information Selector,
        fonttitle=\small\bfseries,
        colbacktitle=white,
        coltitle=black,
        colback=white,
        colframe=black!80,
        arc=1mm,
        boxrule=0.5pt,
        boxsep=1pt,
        toptitle=1pt,
        bottomtitle=1pt,
        enhanced,
        drop shadow={shadow xshift=0.5mm, shadow yshift=-0.5mm}
    ]
    \begin{itemize}[leftmargin=2em, label=-, nosep]
        \item Critical Section
        \item Relevance Filtering
        \item Priority Rules
    \end{itemize}
    \end{tcolorbox}
};
\node at ($(context_pres.center)+(4.6cm,-0.25cm)$) {
    \begin{tcolorbox}[
        width=4.8cm,
        title=Context Restoration,
        fonttitle=\small\bfseries,
        colbacktitle=white,
        coltitle=black,
        colback=white,
        colframe=black!80,
        arc=1mm,
        boxrule=0.5pt,
        boxsep=1pt,
        toptitle=1pt,
        bottomtitle=1pt,
        enhanced,
        drop shadow={shadow xshift=0.5mm, shadow yshift=-0.5mm}
    ]
    \begin{itemize}[leftmargin=2em, label=-, nosep]
        \item Checkpoint Mgnt.
        \item Context Reconstruction
        \item Restoration Strategy
    \end{itemize}
    \end{tcolorbox}
};

\node[comp_group=green!40, minimum height=2.8cm, below=0.5cm of context_pres] (validation) {};
\node[group_title] at ($(validation.north)-(0,0.3cm)$) {\normalsize Validation Framework};
\node at ($(validation.center)+(-4.6cm,-0.25cm)$) {
    \begin{tcolorbox}[
        width=4.2cm,
        title=Validation Manager,
        fonttitle=\small\bfseries,
        colbacktitle=white,
        coltitle=black,
        colback=white,
        colframe=black!80,
        arc=1mm,
        boxrule=0.5pt,
        boxsep=1pt,
        toptitle=1pt,
        bottomtitle=1pt,
        enhanced,
        drop shadow={shadow xshift=0.5mm, shadow yshift=-0.5mm}
    ]
    \begin{itemize}[leftmargin=2em, label=-, nosep]
        \item Validation Layers
        \item Orchestration
        \item Result Handling
    \end{itemize}
    \end{tcolorbox}
};
\node at ($(validation.center)+(0.0cm,-0.25cm)$) {
    \begin{tcolorbox}[
        width=4.2cm,
        title=~~~~~Validator Registry,
        fonttitle=\small\bfseries,
        colbacktitle=white,
        coltitle=black,
        colback=white,
        colframe=black!80,
        arc=1mm,
        boxrule=0.5pt,
        boxsep=1pt,
        toptitle=1pt,
        bottomtitle=1pt,
        enhanced,
        drop shadow={shadow xshift=0.5mm, shadow yshift=-0.5mm}
    ]
    \begin{itemize}[leftmargin=2em, label=-, nosep]
        \item Intra/Inter Agents
        \item Validation Rules
        \item Domain Knowledge
    \end{itemize}
    \end{tcolorbox}
};
\node at ($(validation.center)+(4.7cm,-0.25cm)$) {
    \begin{tcolorbox}[
        width=4.5cm,
        title=Protocol Handlers,
        fonttitle=\small\bfseries,
        colbacktitle=white,
        coltitle=black,
        colback=white,
        colframe=black!80,
        arc=1mm,
        boxrule=0.5pt,
        boxsep=1pt,
        toptitle=1pt,
        bottomtitle=1pt,
        enhanced,
        drop shadow={shadow xshift=0.5mm, shadow yshift=-0.5mm}
    ]
    \begin{itemize}[leftmargin=2em, label=-, nosep]
        \item Rejection Protocol
        \item Augmentation Protocol
        \item Feedback Protocol
    \end{itemize}
    \end{tcolorbox}
};
\node[comp_group=cyan!25, minimum height=4.4cm, below=0.5cm of validation] (transaction) {};
\node[group_title] at ($(transaction.north)-(0,0.3cm)$) {\normalsize Transaction Framework};

\node at ($(transaction.center)+(-4.5cm,0.72cm)$) {
    \begin{tcolorbox}[
        width=4.0cm,
        title=Saga Coordinator,
        fonttitle=\small\bfseries,
        colbacktitle=white,
        coltitle=black,
        colback=white,
        colframe=black!80,
        arc=1mm,
        boxrule=0.5pt,
        boxsep=1pt,
        toptitle=1pt,
        bottomtitle=1pt,
        enhanced,
        drop shadow={shadow xshift=0.3mm, shadow yshift=-0.3mm}
    ]
    \begin{itemize}[leftmargin=1em, label=-, nosep]
        \item Saga Definition
        \item Execution Control
        \item Failure Detection
    \end{itemize}
    \end{tcolorbox}
};

\node at ($(transaction.center)+(0.0cm,0.70cm)$) {
    \begin{tcolorbox}[
        width=4.0cm,
        title=~~~Transaction Manager,
        fonttitle=\small\bfseries,
        colbacktitle=white,
        coltitle=black,
        colback=white,
        colframe=black!80,
        arc=1mm,
        boxrule=0.5pt,
        boxsep=1pt,
        toptitle=1pt,
        bottomtitle=1pt,
        enhanced,
        drop shadow={shadow xshift=0.3mm, shadow yshift=-0.3mm}
    ]
    \begin{itemize}[leftmargin=1em, label=-, nosep]
        \item Transaction Log
        \item State Management
        \item Version Control
    \end{itemize}
    \end{tcolorbox}
};

\node at ($(transaction.center)+(4.5cm,0.70cm)$) {
    \begin{tcolorbox}[
        width=4.2cm,
        title=Compensation Manager,
        fonttitle=\small\bfseries,
        colbacktitle=white,
        coltitle=black,
        colback=white,
        colframe=black!80,
        arc=1mm,
        boxrule=0.5pt,
        boxsep=1pt,
        toptitle=1pt,
        bottomtitle=1pt,
        enhanced,
        drop shadow={shadow xshift=0.3mm, shadow yshift=-0.3mm}
    ]
    \begin{itemize}[leftmargin=1em, label=-, nosep]
        \item Compensation Reg.
        \item Rollback Orchestration
        \item Recovery Strategies
    \end{itemize}
    \end{tcolorbox}
};
\node at ($(transaction.center)+(-2.5cm,-1.05cm)$) {
    \begin{tcolorbox}[
        width=4.0cm,
        title=Dependency Tracker,
        fonttitle=\small\bfseries,
        colbacktitle=white,
        coltitle=black,
        colback=white,
        colframe=black!80,
        arc=1mm,
        boxrule=0.5pt,
        boxsep=1pt,
        toptitle=1pt,
        bottomtitle=1pt,
        enhanced,
        drop shadow={shadow xshift=0.5mm, shadow yshift=-0.5mm}
    ]
    \begin{itemize}[leftmargin=1em, label=-, nosep]
        \item Dependency Graph
        \item Condition Evaluation
        \item Satisfaction Checking
    \end{itemize}
    \end{tcolorbox}
};
\node at ($(transaction.center)+(2.5cm,-1.05cm)$) {
    \begin{tcolorbox}[
        width=3.8cm,
        title=Critical Op Manager,
        fonttitle=\small\bfseries,
        colbacktitle=white,
        coltitle=black,
        colback=white,
        colframe=black!80,
        arc=1mm,
        boxrule=0.5pt,
        boxsep=1pt,
        toptitle=1pt,
        bottomtitle=1pt,
        enhanced,
        drop shadow={shadow xshift=0.5mm, shadow yshift=-0.5mm}
    ]
    \begin{itemize}[leftmargin=1em, label=-, nosep]
        \item Operation Marking
        \item Tracking Control
        \item Validation Rules
    \end{itemize}
    \end{tcolorbox}
};

\node[layer=gray!10, minimum height=2.1cm, below=0.8cm of sagallm] (langgraph) {};
\node[layer_title] at ($(langgraph.north)-(0,0.4cm)$) {\LARGE LangGraph or Equivalent};

\node[component, scale=0.98] (stategraph) at ($(langgraph.center)+(-5.7cm,-.18)$) {\normalsize StateGraph};
\node[component, scale=0.98] (state_flow) at ($(langgraph.center)+(-1.9cm,-.18)$) {\normalsize State \& Flow};
\node[component, scale=0.98] (node_handlers) at ($(langgraph.center)+(1.9cm,-.18)$) {\normalsize Node Handlers};
\node[component, scale=0.98] (exec_mgmt) at ($(langgraph.center)+(5.7cm,-.18)$) {\normalsize Execution Management};

\draw[arrow] (app_layer.south) -- (sagallm.north);
\draw[arrow] (context_pres.south) -- (validation.north);
\draw[arrow] (validation.south) -- (transaction.north);
\draw[arrow] (sagallm.south) -- (langgraph.north);

\node[layer=gray!10, minimum height=2.1cm, text width=12.8cm, below=0.8cm of langgraph] (nativellms) {};
\node[layer_title] at ($(nativellms.north)-(0,0.4cm)$) {\LARGE LLM Infrastructure};

\node[component] (languagemodels) at ($(nativellms.center)+(-4.0cm,-.18)$) {\normalsize Language Models (e.g., GPT-4, Claude, etc.) };
\node[component] (llmapis) at ($(nativellms.center)+(0.0cm,-.18)$) {\normalsize Interface Endpoints (APIs)};
\node[component] (contextmanagement) at ($(nativellms.center)+(4.0cm,-.18)$) {\normalsize Context Management (Tokenization, Windows)};

\draw[arrow] (app_layer.south) -- (sagallm.north);
\draw[arrow] (context_pres.south) -- (validation.north);
\draw[arrow] (validation.south) -- (transaction.north);
\draw[arrow] (sagallm.south) -- (langgraph.north);
\draw[arrow] (langgraph.south) -- (nativellms.north);

\end{tikzpicture}
\end{center}
\vspace{-.1in}
\caption{Architecture of $\SAGA$. It sits between the application layer and LLMs, consisting of three frameworks: Context Management, Validation, and Transaction.}
\label{fig:SAGALLM-architecture}
\vspace{-.1in}
\end{figure}

\subsection{Context Management Requirements}

$\SAGA$ identifies and retains essential context for recovery, validation, and inter-agent dependencies:

\begin{itemize}[leftmargin=1.5em, topsep=.05em, parsep=.05em, label=-]
\item \textit{Selective Retention:} Filters critical info
\item \textit{Structured Storage:} Organizes specs, justifications, and reasoning
\item \textit{Dependency Tracking:} Maintains prerequisites for rollback
\item \textit{Communication Protocol:} Ensures necessary context exchange
\end{itemize}

\subsubsection*{Failure Handling and Recovery}

Effective recovery depends on preserved context and dependency tracking. $\SAGA$ supports multi-level failure response:

\begin{enumerate}[leftmargin=1.5em, topsep=.05em, parsep=.05em, label=\arabic*.]
\item \textit{Operation-Level:} Upon failure, the system invokes compensatory actions using logs and rollback specifications stored in $S_O$.
\item \textit{Workflow-Level:} $\SAGA$ traverses the dependency graph to orchestrate reverse execution paths across agents, restoring global consistency based on $S_D$ and recorded constraints.
\end{enumerate}

This layered integration of recovery within strategic context management enables $\SAGA$ to meet the transactional demands of complex real-world multi-agent LLM workflows.

\section{Design and Implementation}
\label{sec:design}

Figure~\ref{fig:SAGALLM-architecture} depicts the $\SAGA$ architecture, which sits between the application layer and LLM multi-agent systems like LangGraph. $\SAGA$ comprises three frameworks: \emph{context management}, \emph{validation}, and \emph{transaction}. 
To illustrate the design, we use a travel planning example.

\subsection*{Travel Planning Problem}
\label{sec:CaseStudy}

This example demonstrates how $\SAGA$ automatically manages complex multi-agent LLM workflows for international trip planning with multiple destinations, budget constraints, and transactional booking requirements. The application illustrates the transition from manual planning to automated $\SAGA$-managed execution.

\subsection{Specifications}
\begin{itemize}[leftmargin=1.2em, topsep=-.05em, parsep=-.05em,label=-]
    \item Plan a trip from San Francisco to Berlin and Cologne and then back to San Francisco.
    \item Travel period: June 2025 (flexible within the month)
    \item Budget constraint: \$5,000 total.
    \item Required bookings: flights, hotels, and trains between cities.
    \item Preferences: 
    \begin{itemize}[leftmargin=1em, topsep=-.05em, parsep=-.05em, label=>]
        \item Moderately priced accommodations (3-4 star hotels).
        \item Direct flights when possible.
        \item Train pass to save money.
        \item Flexible scheduling with 4 days in Berlin and 2 in Cologne. 
    \end{itemize}
\end{itemize}

\subsection{Two-Phase Workflow Architecture}

The application workflow consists of two distinct phases with different automation levels:

\textbf{Phase 1 (Manual Planning)}: Human-driven itinerary planning and user validation.

\textbf{Phase 2 (Automated $\SAGA$ Execution)}: Fully automated multi-agent transaction management.

\begin{figure}[t]
\centering
\begin{tikzpicture}[
    box/.style={rectangle, draw, minimum width=7em, minimum height=2.0em, text centered, fill=gray!10,
                drop shadow={shadow xshift=0.7mm, shadow yshift=-0.7mm, opacity=0.3, fill=black}},
    arrow/.style={->, thick}
]
\node[box] (plan) at (0.6,2.8) {Initial Planning};

\draw[draw, fill=green!30, drop shadow={shadow xshift=0.7mm, shadow yshift=-0.7mm, opacity=0.3, fill=black}] 
    (0.6,2.0) -- (1.8,1.2) -- (0.6,0.4) -- (-0.4,1.2) -- cycle;
\node at (0.6,1.2) {\small\begin{tabular}{c}User\\Validation\end{tabular}};

\coordinate (approval) at (0.6,1.2);
\coordinate (approval-north) at (0.6,2.0);
\coordinate (approval-south) at (0.6,0.4);
\coordinate (approval-east) at (1.8,1.2);
\coordinate (approval-west) at (-0.4,1.2);

\node[box] (revision) at (-2.7,1.2) {Revision};
\node[box] (finalize) at (0.3,-0.60) {Finalize Itinerary};

\node[box, fill=cyan!20] (flight1) at (-2.7,-0.60) {Flight to Berlin};
\node[box, fill=cyan!20] (hotel1) at (-2.7,-1.8) {Berlin Hotel};
\node[box, fill=cyan!20] (train) at (0.3,-1.8) {Train to Cologne};
\node[box, fill=cyan!20] (hotel2) at (3.3,-1.8) {Cologne Hotel};
\node[box, fill=cyan!20] (flight2) at (3.3,-0.60) {Return Flight};

\draw[arrow] (plan) -- (approval-north);
\draw[arrow] (approval-south) -- node[left, font=\small] {Yes} ($(finalize) + (0.3, 0.3)$);
\draw[arrow] (approval-west) -- node[above, font=\small] {No} (revision);
\draw[arrow] ($(revision) + (0,0.3)$) to[out=30,in=90] (approval-north);

\draw[arrow, thick, blue] (finalize.west) -- node[above, font=\small\color{blue}] {$\SAGA$ Takeover} (flight1.east);

\draw[arrow] ($(flight1.south)+(-0.4,0)$) -- ($(hotel1.north)+(-0.4,0)$);
\draw[arrow] ($(hotel1.east)+(-0.0,0.3)$) -- ($(train.west)+(-0.0,0.3)$);
\draw[arrow] ($(train.east)+(0,0.3)$) -- ($(hotel2.west)+(0,0.3)$);
\draw[arrow] ($(hotel2.north)+(-0.4,0)$) -- ($(flight2.south)+(-0.4,0)$);

\draw[arrow, dashed, red] ($(hotel1.north)+(0.40,0)$) -- node[font=\small] {auto compensate} ($(flight1.south)+(0.40,0)$);
\draw[arrow, dashed, red] ($(train.west)+(0,-0.2)$) -- node[above, font=\small] {auto comp} ($(hotel1.east)+(0.0,-0.2)$);
\draw[arrow, dashed, red] ($(hotel2.west)+(0.0,-0.2)$) -- node[above, font=\small] {auto comp} ($(train.east)+(0.0,-0.2)$);
\draw[arrow, dashed, red] ($(flight2.south)+(0.4,0)$) -- node[font=\small] {auto compensate} ($(hotel2.north)+(0.4,0)$);

\draw[arrow, dashed, red] (flight1.north) to[out=90,in=270] node[left, font=\small] {error} (revision.south);


\end{tikzpicture}
\caption{Travel Planning Workflow showing transition from manual planning to automated $\SAGA$ execution. Gray boxes represent manual human-driven activities. Cyan boxes represent fully automated $\SAGA$-managed transactions with automatic compensation, validation, and recovery. The blue arrow indicates the handoff point where $\SAGA$ takes complete control.}
\label{fig:travel-workflow}
\vspace{-.2in}
\end{figure}
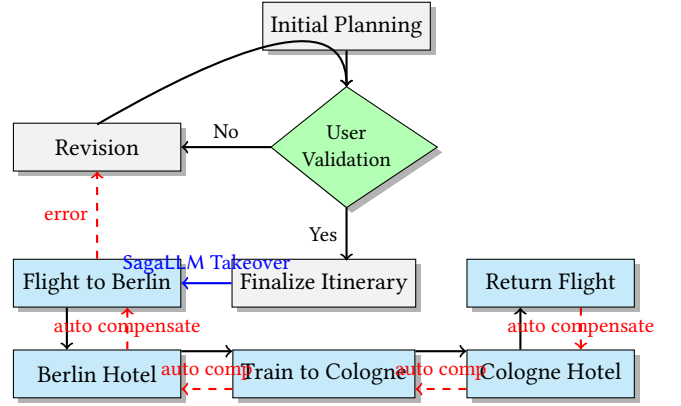

Figure~\ref{fig:travel-workflow} illustrates this workflow transition, where gray boxes represent manual planning activities and cyan boxes represent automated $\SAGA$-managed transactions.

\subsubsection{Phase 1: Manual Itinerary Planning}

Phase 1 involves traditional human-driven planning activities that establish the requirements and constraints for automated execution:

\begin{enumerate}[leftmargin=1.36em, topsep=-.0em, parsep=-.0em, label=\arabic*.] 
    \item \textbf{Initial Plan Generation}: Human planners or basic LLMs generate multiple feasible itineraries based on user requirements, specifying flight options, hotel reservations, and train transportation with cost estimates validated against budget constraints.

    \item \textbf{Iterative Refinement}: Users review itineraries and provide feedback, leading to plan adjustments without automated transaction management. Essential context (dates, preferences, constraints) is tracked for handoff to $\SAGA$, with iterations continuing until user satisfaction is reached.

    \item \textbf{Plan Finalization and $\SAGA$ Handoff}: Users select and approve the final itinerary with all requirements and constraints. The system compiles comprehensive specifications including booking dependencies, budget limits, and user preferences, then initiates \textbf{automated handoff to $\SAGA$} where all subsequent design, coding, agent coordination, and execution becomes fully automated.
\end{enumerate}

\subsubsection{Phase 2: Automated $\SAGA$ Execution Overview}

Once Phase 1 is completed, $\SAGA$ automatically takes full control of workflow development and management. Given a planning problem $\mathcal{O}$, constraint set $D$, and performance metrics $\mathcal{M}$, $\SAGA$ generates a complete workflow consisting of nodes and edges, provides specifications for both transaction agents and compensation agents, and conducts validation and refinement. Each workflow node and edge is assigned both a regular agent that handles transactions and a compensation agent that handles rollback execution.

Algorithm~\ref{alg:ALAS-Meta-Full} outlines $\SAGA$'s complete Phase 2 workflow and code generation process. For in-depth discussion on code development and run-time monitoring, please refer to the extended version of this work \cite{chang2025ALAS}.


\begin{algorithm*}[ht!]
\caption{Workflow $\mathcal{W_\text{template}}$ Construction and Agent Code Generation}
\label{alg:ALAS-Meta-Full}
\begin{small}
\begin{multicols}{2}
\setlength{\multicolsep}{2pt}
\begin{algorithmic}[1]
\Require Problem specification $\mathcal{O}$, constraints $D$, performance metrics $\mathcal{M}$
\Statex \hspace{-1.9em} \textbf{Local Variables:}
\State Roles $\mathcal{R}$; Profiles $\mathcal{P}$; Nodes $\mathcal{N}$; Edges $\mathcal{E}$
\State Log schemas $\mathcal{L}_{n_i}, \mathcal{L}_{e_{ij}}$
\State Agents, Comp Agents $\alpha_{n_i}, \alpha_{e_{ij}}$, $\alpha^{comp}_{n_i}, \alpha^{comp}_{e_{ij}}$
\Ensure Validated $\mathcal{W_\text{template}} = (\mathcal{N}, \mathcal{E})$
\Statex

\Statex \hspace{-1.5em} \textbf{Stage 1: Network Construction} (extracting information from $\mathcal{O}$)
\State $\mathcal{R} \gets \text{ExtractRoles}(\mathcal{O})$
\State $\{(n_i, \mathcal{P}_i)\} \gets \text{map}_{\text{role}}(\mathcal{O}, \mathcal{R})$ 
\State $\mathcal{N} \gets \{n_i\}$, $\mathcal{E} \gets \text{map}_{\text{dep}}(\mathcal{N}, D)$ 
\State $\mathcal{W_\text{template}} \gets (\mathcal{N}, \mathcal{E})$
\Statex

\Statex \hspace{-1.5em} \textbf{Stage 2: Agent Specification} (for each node and edge, creating its agent and compensation agent with logging schema)
\ForAll{$n_i \in \mathcal{N}$}
  \State $\mathcal{L}_{n_i} \gets \text{DefineLogSchema}(n_i, \mathcal{P}_{n_i})$
  \State $\alpha_{n_i} \gets \text{DefineNodeAgent}(n_i, \mathcal{L}_{n_i})$
  \State $\alpha^{comp}_{n_i} \gets \text{DefineCompAgent}(\alpha_{n_i}, \mathcal{L}_{n_i})$
\EndFor
\ForAll{$e_{ij} \in \mathcal{E}$}
  \State $\mathcal{L}_{e_{ij}} \gets \text{DefineLogSchema}(e_{ij}, \mathcal{P}_{e_{ij}})$
  \State $\alpha_{e_{ij}} \gets \text{DefineEdgeAgent}(e_{ij}, \mathcal{L}_{e_{ij}})$
  \State $\alpha^{comp}_{e_{ij}} \gets \text{DefineCompAgent}(\alpha_{e_{ij}}, \mathcal{L}_{e_{ij}})$
\EndFor
\Statex

\Statex \hspace{-1.5em} \textbf{Stage 3: Validation and Refinement} (code validation and refinement)
\State $\mathcal{W_\text{template}} \gets \text{UpdateWorkflow}(\mathcal{N}, \mathcal{E}, \alpha, \alpha^{comp})$
\While{not $\text{ValidateWorkflow}(\mathcal{W_\text{template}}, \mathcal{M})$}
  \State $\text{StructuralValidation}(\mathcal{W_\text{template}})$
  \State $\text{ConstraintValidation}(\mathcal{W_\text{template}}, D)$
  \State $\text{CompensationValidation}(\mathcal{W_\text{template}}), \{\alpha^{comp}\})$
  \State $\mathcal{W_\text{template}} \gets \text{RefineWorkflow}(\mathcal{W_\text{template}}, \mathcal{M})$
\EndWhile
\State \Return $\mathcal{W_\text{template}}$
\end{algorithmic}
\end{multicols}
\end{small}
\vspace{-0.1in}
\end{algorithm*}

\subsubsection{Detailed Phase 2 Implementation}

The automated execution phase consists of four integrated components:

\begin{enumerate}[leftmargin=1.36em, topsep=.05em, parsep=.05em, label=\arabic*.] 
    \item \textbf{Automatic System Architecture Generation}: $\SAGA$ analyzes the finalized itinerary and automatically generates the appropriate agent architecture, defines the transaction sequences ($T_1, T_2, ..., T_n$) and the corresponding compensations ($C_1, C_2, ..., C_n$), and establishes validation rules and dependency graphs based on booking requirements.

    \item \textbf{Automatic Agent Deployment and Coordination}: The system instantiates required domain agents (FlightBookingAgent, HotelBookingAgent, etc.) with appropriate configurations, deploys GlobalValidationAgent and SagaCoordinatorAgent with full system access, and configures agent communication protocols and data schemas.

    \item \textbf{Automatic Transaction Execution}: The system executes the complete booking sequence: $T_1$ (International Flight Booking SFO $\rightarrow$ Berlin), $T_2$ (Berlin Hotel Booking coordinated with flight confirmation), $T_3$ (Train Booking Berlin $\rightarrow$ Cologne scheduled with hotel checkout), $T_4$ (Cologne Hotel Booking aligned with train arrival), and $T_5$ (International Return Flight Cologne $\rightarrow$ SFO coordinated with hotel checkout).

    \item \textbf{Automatic Exception Handling and Recovery}: $\SAGA$ automatically detects validation failures and executes appropriate compensations, maintains system consistency without human intervention, automatically replans affected portions using preserved context and constraints, and only falls back to Phase 1 for human re-evaluation when automatic replanning cannot satisfy constraints.
\end{enumerate}

\subsection{Agent Architecture and Code Structures}

\begin{figure*}[t!]
\vspace{-.1in}
\centering
\begin{tikzpicture}[
    level 1/.style={sibling distance=55mm, level distance=10mm},
    level 2/.style={sibling distance=17.5mm, level distance=10mm},
    level 3/.style={sibling distance=17.5mm, level distance=10.5mm},
    edge from parent/.style={draw, -latex},
    every node/.style={draw, rounded corners, fill=cyan!20, text width=4.45em, align=center, font=\footnotesize, 
                       drop shadow={shadow xshift=0.7mm, shadow yshift=-0.7mm, opacity=0.3, fill=black}},
    file/.style={fill=blue!1, drop shadow={shadow xshift=0.5mm, shadow yshift=-0.5mm, opacity=0.2, fill=black}}
]
    \node[fill=white!10] (root) {Root}
        child {node[fill=gray!10] (integration) {langgraph\_\\integration}
            child {node[file] {saga\_\\graph.py}}
            child {node[file] {state\_\\manager.py}}
            child {node[file] {agent\_\\wrappers.py}}
        }
        child {node (sagallm) {$\SAGA$}
            child {node (context) {context management}}
            child {node[fill=green!40] (validation) {agent validation}}
            child {node (transaction) {transaction management}}
        }
        child {node[fill=gray!20] (applications) {applications}
            child {node[file] {travel\_\\planning.py}}
            child {node[file] {customer\_\\support.py}}
            child {node[fill=gray!20] (complex) {complex}
                child {node[file] {multi\_agent\_\\workflow.py}}
                child {node[file] {nested\_tran-\\sactions.py}}
            }
        };
    
    \coordinate (contextx) at (context);
    \coordinate (validationx) at (validation);
    \coordinate (transactionx) at (transaction);
    
    \node[file] (state) at (-3.2,-3.8) {app\_state.py};
    \node[file] (RAG) at (-5.0,-3.8) {RAG.py};
    \node[file] (contextS) at (-5.0,-3.0) {select\_\\context.py};
    \node[file] (contextR) at (-3.2,-3.0) {restore\_\\context.py};
    
    \node[file] (intra) at (-1.2,-3) {intra\_\\agent.py};
    \node[file] (validators) at (-1.2,-3.8) {validators.py};
    \node[file] (inter) at (0.5,-3) {inter\_\\agent.py};
    \node[file] (protocols) at (0.5,-3.8) {protocols.py};
    
    \node[file] (manager) at (2.5,-3.0) {transaction\_\\manager.py};
    \node[file] (saga) at (4.2,-3.0) {saga\_\\coordinator.py};
    \node[file] (compensation) at (4.2,-3.8) {compensation};
    \node[file] (dependency) at (2.5,-3.8) {dependency.py};
    
    \draw[-latex] (context) -- (contextS);
    \draw[-latex] (context) -- (contextR);
    
    \draw[-latex] (validation) -- (intra);
    \draw[-latex] (validation) -- (inter);
    
    \draw[-latex] (transaction) -- (saga);
    \draw[-latex] (transaction) -- (manager);
    
\end{tikzpicture}
\caption{Directory Structure of $\SAGA$ Integration with LangGraph}
\label{fig:directory-structure}
\end{figure*}
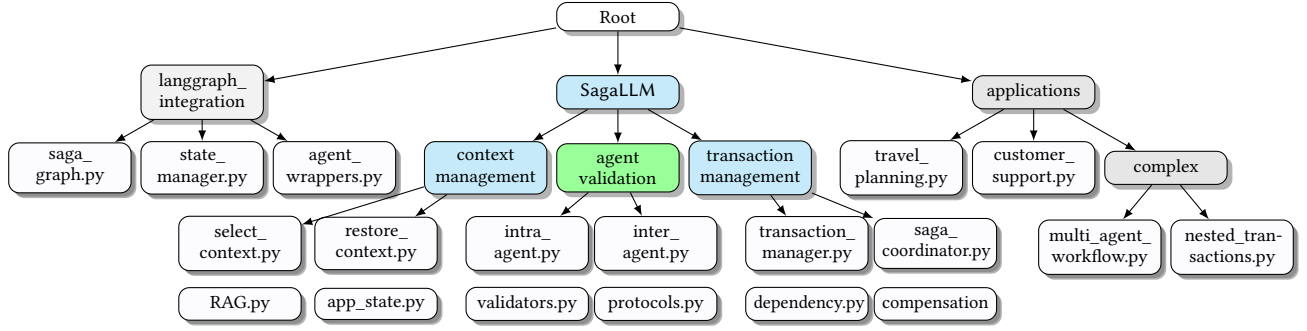

Figure~\ref{fig:directory-structure} presents the organization of the code implementation, structured into the application interface, $\SAGA$ core, and LangGraph integration components. The agent architecture consists of task execution agents that handle domain-specific operations and global coordination agents that manage system-wide consistency and validation.


\subsubsection{Task Execution Agents}

Task execution agents focus solely on their domain-specific operations, with all validation handled externally by the global validation agent. Each agent maintains structured input/output interfaces and internal state for compensation purposes.

\paragraph{\textbf{FlightBookingAgent}}
\begin{itemize}[leftmargin=1em, topsep=.05em, parsep=.05em, label=-]
    \item \textit{Input Schema}: {travel\_dates, budget\_limit, airline\_preferences, passenger\_details}
    \item \textit{Output Schema}: {flight\_details, confirmation\_number, total\_cost, cancellation\_policy}
    \item \textit{Internal State}: {reservation\_status, booking\_reference, payment}
\end{itemize}

\paragraph{\textbf{HotelBookingAgent}}
\begin{itemize}[leftmargin=1em, topsep=.05em, parsep=.05em, label=-]
    \item \textit{Input Schema}: {checkin\_date, checkout\_date, location\_constraints, amenity\_preferences, budget\_limit}
    \item \textit{Output Schema}: {hotel\_details, room\_type, confirmation\_number, total\_cost, cancellation\_policy}
    \item \textit{Internal State}: {reservation\_status, booking\_reference, payment}
\end{itemize}

\paragraph{\textbf{TrainBookingAgent}}
\begin{itemize}[leftmargin=1em, topsep=.05em, parsep=.05em, label=-]
    \item \textit{Input Schema}: {departure\_location, arrival\_location, travel\_time, connection\_requirements}
    \item \textit{Output Schema}: {train\_details, seat\_res, total\_cost, schedule\_details}
    \item \textit{Internal State}: {ticket\_status, booking\_reference, refund\_policy}
\end{itemize}

\paragraph{\textbf{BudgetTrackingAgent}}
\begin{itemize}[leftmargin=1em, topsep=.05em, parsep=.05em, label=-]
    \item \textit{Input Schema}: {expense\_item, cost, category, transaction\_id}
    \item \textit{Output Schema}: {updated\_total, remaining\_budget, budget\_status, expense\_breakdown}
    \item \textit{Internal State}: {cumulative\_expenses, expense\_log, constraints}
\end{itemize}

\paragraph{\textbf{ItineraryPlanningAgent}}
\begin{itemize}[leftmargin=1em, topsep=-.05em, parsep=-.05em, label=-]
    \item \textit{Input Schema}: {user\_prefs, travel\_constraints, confirmations}
    \item \textit{Output Schema}: {optimized\_itinerary, timing\_schedule, activity recommendations}
    \item \textit{Internal State}: {preference\_history, optimization\_parameters, constraint\_violations}
\end{itemize}

\subsubsection{Global Coordination Agents}
\label{sec:globalCoordination}

\paragraph{\textbf{GlobalValidationAgent}}
The central validation authority that maintains access to all system state and performs comprehensive validation before any transaction commitment.
\begin{itemize}[leftmargin=1em, topsep=-.05em, parsep=-.05em, label=-]
    \item \textit{System Access}: Complete visibility to all agent outputs, transaction history, dependency graph, and critical context
    \item \textit{Validation Scope}: Intra-agent output validation and inter-agent communication validation (detailed in Table~\ref{tab:travelvalidation})
    \item \textit{Response Protocols}: Rejection (triggers compensation), Augmentation (enhances outputs), Feedback (improves future performance)
\end{itemize}

\paragraph{\textbf{SagaCoordinatorAgent}}
Manages transaction sequencing, dependency tracking, and compensation orchestration.
\begin{itemize}[leftmargin=1em, topsep=-.05em, parsep=-.05em, label=-]
    \item \textit{Coordination State}: {active\_transactions, dependency\_graph, compensation\_queue, transaction\_log}
    \item \textit{Responsibilities}: Transaction ordering, failure detection, compensation sequence execution
\end{itemize}

\subsubsection{Critical Context and State Management}
\label{sec:contextmanagement}

$\SAGA$ maintains comprehensive context across three state dimensions, with specific agents responsible for different aspects:

\paragraph{\textbf{Application State ($S_A$)}} Managed by task execution agents:
\begin{itemize}[leftmargin=1em, topsep=-.05em, parsep=-.05em, label=-]
    \item \textit{Travel Configuration}: {travel\_dates\_per\_city, destination\_sequence, passenger\_manifest}
    \item \textit{Booking Details}: {confirmation\_numbers, cancellation\_policies, pricing\_breakdown}
    \item \textit{User Constraints}: {budget\_limits, preference\_profiles, accessibility\_requirements}
\end{itemize}

\paragraph{\textbf{Operation State ($S_O$)}} Managed by SagaCoordinatorAgent:
\begin{itemize}[leftmargin=1em, topsep=-.05em, parsep=-.05em, label=-]
    \item \textit{Transaction Log}: {transaction\_id, agent\_id, input\_data, output\_data, execution\_timestamp}
    \item \textit{Decision Reasoning}: {reasoning\_chain, alternatives\_considered, decision\_justification}
    \item \textit{Compensation Actions}: {compensation\_procedure, rollback requirements, recovery\_state}
\end{itemize}

\paragraph{\textbf{Dependency State ($S_D$)}} Managed by GlobalValidationAgent:
\begin{itemize}[leftmargin=1em, topsep=-.05em, parsep=-.05em, label=-]
    \item \textit{Inter-Booking Dependencies}: {prerequisite\_transactions, temporal\_constraints, resource\_dependencies}
    \item \textit{Validation Status}: {validation\_results, constraint\_satisfaction, dependency\_resolution}
\end{itemize}

\subsection{Transaction Flow and Validation Protocol}

\subsubsection{Transaction Execution Sequence}

Each transaction follows a standardized execution pattern managed by SagaCoordinatorAgent with validation checkpoints enforced by  GlobalValidationAgent:

\begin{enumerate}[leftmargin=1.36em, topsep=-.05em, parsep=-.05em, label=\arabic*.]
    \item \textit{Pre-execution Validation}: GlobalValidationAgent validates inputs and dependency satisfaction
    \item \textit{Transaction Execution}: Task agent performs operation
    \item \textit{Output Validation}: GlobalValidationAgent performs comprehensive output validation (Table~\ref{tab:travelvalidation})
    \item \textit{State Commitment}: Upon validation success, results are committed to system state
    \item \textit{Compensation Registration}: SagaCoordinatorAgent records compensation procedures for potential rollback
\end{enumerate}

\subsubsection{Comprehensive Validation Framework}

Table~\ref{tab:travelvalidation} details the validation types performed by the GlobalValidationAgent at each checkpoint. All validation occurs externally to task agents, ensuring independent eval of agent outputs and inter-agent communications.

\begin{table}[t]
\centering
\footnotesize
\caption{Validations Performed by GlobalValidationAgent}
\vspace{-.1in}
\label{tab:travelvalidation}
\begin{tabular}{>{\raggedright\arraybackslash}p{2.5cm}>{\raggedright\arraybackslash}p{4.5cm}}
\toprule
\textbf{Validation Type} & \textbf{Implementation Example} \\
\midrule
\multicolumn{2}{l}{\textit{\textbf{Intra-Agent Output Validation}}} \\
\midrule
Syntactic Validation & Verify JSON structure with required fields (departure\_time, arrival\_time, flight\_number) \\
\midrule
Semantic Validation & Confirm accommodation covers entire trip duration without gaps \\
\midrule
Factual Validation & Maintain consistent travel times (45-minute hotel-to-train travel time) \\
\midrule
Constraint Adherence & Enforce budget limits (total cost under \$5,000 maximum) \\
\midrule
Reasoning Validation & Verify logical decision chains (weather-based activity recommendations) \\
\midrule
\multicolumn{2}{l}{\textit{\textbf{Inter-Agent Communication Validation}}} \\
\midrule
Dependency Satisfaction & Ensure flight booking completion before hotel finalization \\
\midrule
Consistency Checks & Standardize location data formats across all agents \\
\midrule
Temporal Validation & Sequence budget finalization after all booking verifications \\
\midrule
Mutual Agreement & Coordinate feasible travel times between transportation and itinerary agents \\
\midrule
Transaction Boundary Integrity & Trigger compensation cascade when flight booking fails \\
\bottomrule
\end{tabular}
\vspace{-.1in}
\end{table}

\subsection{Compensation and Recovery Mechanisms}

\subsubsection{Transaction-Specific Compensations}

Each transaction maintains explicit compensation procedures executed by the SagaCoordinatorAgent upon validation failure:

\paragraph{\textbf{Flight Booking Compensation ($C_1$)}}
\begin{itemize}[leftmargin=1em, topsep=-.05em, parsep=-.05em, label=-]
    \item \textit{Immediate Actions}: Cancel reservation, release seat, refund
    \item \textit{State Restoration}: Reset booking status, clear confirmation numbers, restore budget allocation
    \item \textit{Dependency Impact}: Trigger hotel and train booking re-evaluation based on new flight availability
\end{itemize}

\paragraph{\textbf{Hotel Booking Compensation ($C_2$)}}
\begin{itemize}[leftmargin=1em, topsep=-.05em, parsep=-.05em, label=-]
    \item \textit{Immediate Actions}: Cancel reservation per hotel policy, refund
    \item \textit{State Restoration}: Cancel room, restore budget allocation
    \item \textit{Dependency Impact}: Notify itinerary planning for location-based activity adjustments
\end{itemize}

\paragraph{\textbf{Train Booking Compensation ($C_3$)}}
\begin{itemize}[leftmargin=1em, topsep=-.05em, parsep=-.05em, label=-]
    \item \textbf{Immediate Actions}: Cancel tickets per railway policy, refund
    \item \textbf{State Restoration}: Clear reservations, update travel schedule
    \item \textbf{Dependency Impact}: Recalculate inter-city travel times for dependent bookings
\end{itemize}

\subsubsection{Recovery Protocol Execution}

Upon validation failure, the system executes a structured recovery sequence:

\begin{enumerate}[leftmargin=1.36em, topsep=.05em, parsep=.05em, label=\arabic*.]
    \item \textbf{Failure Detection}: GlobalValidationAgent identifies validation failure and triggers compensation
    \item \textbf{Dependency Analysis}: SagaCoordinatorAgent analyzes dependency graph to determine affected transactions
    \item \textbf{Compensation Sequence}: Execute compensations in reverse dependency order ($C_n, C_{n-1}, ..., C_1$)
    \item \textbf{State Verification}: GlobalValidationAgent confirms system state consistency after compensation
    \item \textbf{Replanning Initiation}: Re-execute affected portion of workflow with preserved context and constraints
\end{enumerate}

This integrated architecture ensures that $\SAGA$'s sophisticated validation, state management, and context preservation requirements are systematically implemented through clear agent responsibilities and structured coordination protocols.

\section{Experiments}
\label{sec:exp}

Having established the theoretical foundation for $\SAGA$ and demonstrated its architecture through the travel planning case study, we now empirically validate the framework's effectiveness in addressing the fundamental limitations of current multi-LLM agent systems. As identified in Section~\ref{sec:related}, existing frameworks suffer from four critical shortcomings that prevent reliable deployment in complex, real-world scenarios: inadequate self-validation capabilities stemming from inherent LLM limitations, context narrowing that leads to information loss during extended workflows, lack of transactional properties that compromise consistency and recovery, and insufficient inter-agent coordination that results in workflow fragmentation.

The $\SAGA$ framework addresses these limitations through its three core requirements established in Section~\ref{sec:requirements}: independent validation to overcome self-validation gaps, automatic context preservation to maintain critical information throughout extended interactions, and comprehensive transactional integrity to ensure consistent state management and reliable recovery. The travel planning specification in the previous section illustrates how these requirements translate into practical system architecture, where manual planning (Phase 1) hands off to fully automated $\SAGA$ execution (Phase 2) with complete transaction management.

Our experiments aim to validate $\SAGA$'s effectiveness by measuring improvements in each of the four problematic areas.
\begin{enumerate}[leftmargin=1.36em, topsep=-.05em, parsep=-.05em, label=\arabic*.]
    \item We aim to compare $\SAGA$-managed workflows against baseline multi-agent systems to demonstrate quantifiable improvements in i) validation accuracy, ii) context retention, iii) transactional consistency, and iv) coordination reliability across diverse planning scenarios.
    \item We seek to demonstrate that $\SAGA$ can handle unexpected plan disruptions and automatically conduct effective reactive planning, combining the strengths of traditional distributed-MAS and LLM-based MAS while mitigating their limitations.
\end{enumerate}

\subsection{Experimental Design}

We selected test cases from the REALM benchmark \cite{2025realmbench}, which evaluates multi-agent systems on distinct problems spanning various complexity levels and coordination requirements. For our experiments, we focused on two medium-tier sequential planning challenges (problems \#5 and \#6) that test systematic workflow execution and dependency management, and two reactive planning challenges (problems \#8 and \#9) that evaluate dynamic adaptation and compensation capabilities. These problems provide comprehensive coverage of the scenarios where the four identified shortcomings most significantly impact system performance.


We evaluated four LLMs---Claude~3.7 \cite{anthropic2024claude}, DeepSeek~R1 \cite{deepseekai2025deepseekr1}, GPT-4o \cite{openai2024gpt4o}, and GPT-o1---alongside our proposed $\SAGA$ framework. All experiments were conducted between March~12 and~17, 2025.
The source code of $\SAGA$ for conducting these experiments is available at \cite{2025SagaCodePart1}. 

\begin{table}[t!]
\centering
\caption{Thanksgiving Dinner Coordination Problem}
\label{tab:ThanksgivingDinner1}
\begin{small}
\renewcommand{\arraystretch}{1.1}
\fbox{
\begin{minipage}{0.45\textwidth}
\textbf{Objective:} Coordinate family arrivals and dinner preparation for 6:00 PM dinner in Boston

\textbf{Family Members and Arrivals:}
\begin{itemize}[leftmargin=1em, topsep=-.1pt, itemsep=-.1pt, label=-]
\item Sarah (Mom): Host, at home
\item James (Dad): Lands at BOS 1:00 PM from SF
\item Emily (Sister): Lands at BOS 2:30 PM from Chicago
\item Michael (Brother): Driving, arrives 3:00 PM from NY
\item Grandma: Needs pickup from suburban Boston
\end{itemize}

\textbf{Cooking Requirements:}
\begin{itemize}[leftmargin=1em, topsep=-.1pt, itemsep=-.1pt, label=-]
\item Turkey: 4 hours cooking time
\item Side dishes: 2 hours preparation
\item Someone must stay home during cooking for fire safety
\end{itemize}

\textbf{Transportation Constraints:}
\begin{itemize}[leftmargin=1em, topsep=-.1pt, itemsep=-.1pt, label=-]
\item James must rent car after landing
\item Emily requires airport pickup
\item Travel times:
   \begin{itemize}
   \item Home to BOS Airport: 60 min
   \item BOS Airport to Grandma's: 60 min
   \item Home to Grandma's: 30 min
   \end{itemize}
\end{itemize}

\textbf{Key Requirements:}
\begin{itemize}[leftmargin=1em, topsep=-.1pt, itemsep=-.1pt, label=-]
\item All family members at home for 6:00 PM dinner
\item Turkey and sides ready by dinner time
\item All pickups completed with available drivers
\item Cooking supervision maintained
\end{itemize}
\end{minipage}
}
\end{small}
\vspace{-.15in}
\end{table}

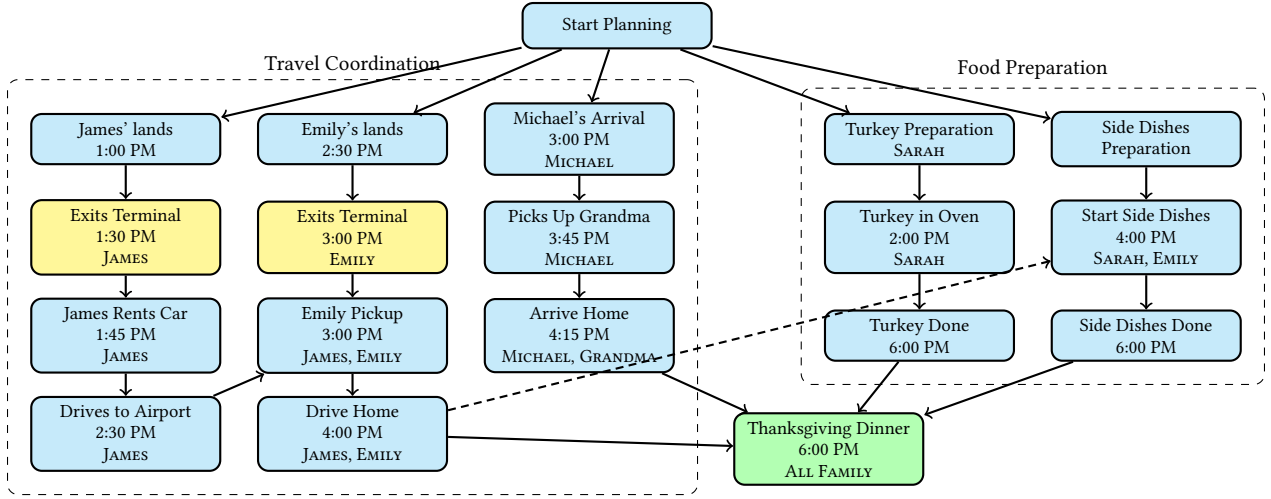
\begin{figure*}[t!]
\vspace{-.1in}
\centering
\begin{tikzpicture}[
    node distance=0.65cm and 1.6cm,
    auto,
    thick,
    event/.style={draw, rounded corners, fill=cyan!20, minimum width=2.5cm, minimum height=0.6cm, font=\footnotesize, align=center},
    critical/.style={draw, rounded corners, fill=yellow!50, minimum width=2.5cm, minimum height=0.6cm, font=\footnotesize, align=center},
    milestone/.style={draw, rounded corners, fill=green!30, minimum width=2.5cm, minimum height=0.6cm, font=\footnotesize, align=center},
    touchdown/.style={draw, rounded corners, fill=green!30, minimum width=2.5cm, minimum height=0.6cm, font=\footnotesize, align=center},
    person/.style={font=\scriptsize\bfseries, text=blue},
    group/.style={draw, dashed, rounded corners, inner sep=0.3cm}
]

\pgfmathsetmacro{\figwidth}{16cm}

\node[event] (start) at (0, +0.2) {Start Planning};

\node[event] (james) at (-6.5,-1.3) {James' lands\\1:00 PM};
\node[critical] (jamesExit) at (-6.5,-2.6) {Exits Terminal\\1:30 PM\\{\textsc{James}}};
\node[event] (rent) at (-6.5,-3.9) {James Rents Car\\1:45 PM\\{\textsc{James}}};
\node[event] (drive) at (-6.5,-5.2) {Drives to Airport\\2:30 PM\\{\textsc{James}}};

\node[event] (emily) at (-3.5,-1.3) {Emily's lands\\2:30 PM};
\node[critical] (emilyExit) at (-3.5,-2.6) {Exits Terminal\\3:00 PM\\{\textsc{Emily}}};
\node[event] (pickup) at (-3.5,-3.9) {Emily Pickup\\3:00 PM\\{\textsc{James, Emily}}};
\node[event] (return) at (-3.5,-5.2) {Drive Home\\4:00 PM\\{\textsc{James, Emily}}};

\node[event] (michael) at (-0.5,-1.3) {Michael's Arrival\\3:00 PM\\{\textsc{Michael}}};
\node[event] (grandma) at (-0.5,-2.6) {Picks Up Grandma\\3:45 PM\\{\textsc{Michael}}};
\node[event] (grandmaReturn) at (-0.5,-3.9) {Arrive Home\\4:15 PM\\{\textsc{Michael, Grandma}}};

\node[event] (turkeyPrep) at (4,-1.3) {Turkey Preparation\\{\textsc{Sarah}}};
\node[event] (turkeyOven) at (4,-2.6) {Turkey in Oven\\2:00 PM\\{\textsc{Sarah}}};
\node[event] (turkeyDone) at (4,-3.9) {Turkey Done\\6:00 PM};

\node[event] (sidePrep) at (7,-1.3) {Side Dishes\\Preparation};
\node[event] (sideStart) at (7,-2.6) {Start Side Dishes\\4:00 PM\\{\textsc{Sarah, Emily}}};
\node[event] (sideDone) at (7,-3.9) {Side Dishes Done\\6:00 PM};

\node[touchdown] (dinner) at (2.8,-5.4) {Thanksgiving Dinner\\6:00 PM\\{\textsc{All Family}}};

\draw[->] (start) -- (james);
\draw[->] (start) -- (emily);
\draw[->] (start) -- (michael);
\draw[->] (start) -- (turkeyPrep);
\draw[->] (start) -- (sidePrep);

\draw[->] (james) -- (jamesExit);
\draw[->] (jamesExit) -- (rent);
\draw[->] (rent) -- (drive);
\draw[->] (drive) -- (pickup);

\draw[->] (emily) -- (emilyExit);
\draw[->] (emilyExit) -- (pickup);
\draw[->] (pickup) -- (return);

\draw[->] (michael) -- (grandma);
\draw[->] (grandma) -- (grandmaReturn);

\draw[->] (turkeyPrep) -- (turkeyOven);
\draw[->] (turkeyOven) -- (turkeyDone);
\draw[->] (turkeyDone) -- (dinner);

\draw[->] (sidePrep) -- (sideStart);
\draw[->] (sideStart) -- (sideDone);
\draw[->] (sideDone) -- (dinner);

\draw[dash pattern=on 3pt off 2pt, ->] (return) -- (sideStart);
\draw[->] (return) -- (dinner);
\draw[->] (grandmaReturn) -- (dinner);

\begin{pgfonlayer}{background}
    \node[group, fit=(james) (jamesExit) (rent) (drive) (emily) (emilyExit) (pickup) (return) (michael) (grandma) (grandmaReturn)] (travel) {};
    \node[font=\small, above] at (travel.north) {Travel Coordination};
    
    \node[group, fit=(turkeyPrep) (turkeyOven) (turkeyDone) (sidePrep) (sideStart) (sideDone)] (food) {};
    \node[font=\small, above] at (food.north) {Food Preparation};
\end{pgfonlayer}

\end{tikzpicture}
\caption{Thanksgiving Dinner Planning Workflow with Common Sense Augmentation, Generated by Claude 3.7}
\label{fig:thanksgiving-workflow-p6}
\end{figure*}

\subsection{Thanksgiving Dinner Problems: P6 and P9}
\label{sec:exp-p6p9}

\begin{figure*}[ht!]
\centering
\begin{tikzpicture}[
    node distance=0.65cm and 1.6cm,
    auto,
    thick,
    event/.style={draw, rounded corners, fill=cyan!20, minimum width=2.4cm, minimum height=0.6cm, font=\footnotesize, align=center},
    critical/.style={draw, rounded corners, fill=red!50, minimum width=2.4cm, minimum height=0.6cm, font=\footnotesize, align=center},
    milestone/.style={draw, rounded corners, fill=green!30, minimum width=2.4cm, minimum height=0.6cm, font=\footnotesize, align=center},
    cancelled/.style={draw, rounded corners, fill=gray!30, minimum width=2.4cm, minimum height=0.6cm, font=\footnotesize, align=center, text=gray},
    alert/.style={draw, rounded corners, fill=red!30, minimum width=2.4cm, minimum height=0.6cm, font=\footnotesize, align=center},
    updated/.style={draw, rounded corners, fill=yellow!60, minimum width=2.4cm, minimum height=0.6cm, font=\footnotesize, align=center},
    person/.style={font=\scriptsize\bfseries, text=blue},
    group/.style={draw, dashed, rounded corners, inner sep=0.3cm}
]

\node[cancelled] (james) at (-6.5,-2.5) {James' Original\\Arrival: 1:00 PM\\{\textsc{CANCELLED}}};
\node[cancelled] (jamesExit) at (-6.5,-3.8) {Original Exit Plan\\{\textsc{CANCELLED}}};
\node[cancelled] (rent) at (-6.5,-5.1) {Original Car Rental\\{\textsc{CANCELLED}}};

\node[updated] (jamesNew) at (-6.5,-6.4) {James' New Arrival\\4:00 PM};
\node[event] (jamesExitNew) at (-6.5,-7.7) {James Exits\\Terminal\\4:30 PM};
\node[event] (rentNew) at (-3.5,-7.7) {James Rents Car\\4:45 PM\\{\textsc{James}}};

\node[event] (emily) at (-3.5,-2.5) {Emily's Arrival\\2:30 PM};
\node[event] (emilyExit) at (-3.5,-3.8) {Emily Exits\\Terminal\\3:00 PM\\{\textsc{Emily}}};
\node[updated] (pickupNew) at (-3.5,-5.1) {Emily Pickup\\3:15 PM\\{\textsc{Sarah}}};
\node[event] (returnNew) at (-3.5,-6.4) {Drive Home\\4:15 PM\\{\textsc{Sarah, Emily}}};

\node[event] (michael) at (-0.5,-2.5) {Michael's Arrival\\3:00 PM\\{\textsc{Michael}}};
\node[event] (grandma) at (-0.5,-3.8) {Picks Up Grandma\\3:45 PM\\{\textsc{Michael}}};
\node[event] (grandmaReturn) at (-0.5,-5.1) {Arrive Home\\4:15 PM\\{\textsc{Michael, Grandma}}};

\node[event] (turkeyPrep) at (4,-2.5) {Turkey Preparation\\{\textsc{Sarah}}};
\node[event] (turkeyOven) at (4,-3.8) {Turkey in Oven\\2:00 PM\\{\textsc{Sarah}}};
\node[alert] (sarahLeaves) at (4,-5.1) {Sarah Leaves for\\Airport\\2:45 PM};
\node[event] (turkeyDone) at (4,-6.4) {Turkey Done\\6:00 PM};

\node[event] (sidePrep) at (7,-2.5) {Side Dishes\\Preparation};
\node[alert] (sideStart) at (7,-5.1) {Start Side Dishes\\4:30 PM\\{\textsc{Emily, Sarah}}};
\node[alert] (sideDone) at (7,-6.4) {Side Dishes Done\\6:00 PM};

\node[alert] (dinner) at (2.8,-7.9) {Thanksgiving Dinner\\{\textbf{\color{red}6:30 PM}}\\{\textsc{All Family}}};

\node[event] (jamesHome) at (-0.5,-7.7) {James Arrives Home\\5:30 PM\\{\textsc{James}}};

\draw[->, dashed, gray] (james) -- (jamesExit);
\draw[->, dashed, gray] (jamesExit) -- (rent);

\draw[->] (jamesNew) -- (jamesExitNew);
\draw[->] (jamesExitNew) -- (rentNew);
\draw[->] (rentNew) -- (jamesHome);
\draw[->] (jamesHome) -- (dinner);

\draw[->] (emily) -- (emilyExit);
\draw[->] (emilyExit) -- (pickupNew);
\draw[->] (pickupNew) -- (returnNew);
\draw[dash pattern=on 3pt off 2pt, ->] (returnNew) -- (sideStart);
\draw[->] (returnNew) -- (dinner);

\draw[->] (michael) -- (grandma);
\draw[->] (grandma) -- (grandmaReturn);
\draw[->] (grandmaReturn) -- (dinner);

\draw[->] (turkeyPrep) -- (turkeyOven);
\draw[->] (turkeyOven) -- (sarahLeaves);
\draw[->] (sarahLeaves) -- (pickupNew);
\draw[->] (turkeyOven) -- (turkeyDone);
\draw[->] (turkeyDone) -- (dinner);

\draw[->] (sidePrep) -- (sideStart);
\draw[->] (sideStart) -- (sideDone);
\draw[->] (sideDone) -- (dinner);

\begin{pgfonlayer}{background}
    \node[group, fit=(james) (jamesExit) (rent) (jamesNew) (jamesExitNew) (rentNew) (emily) (emilyExit) (pickupNew) (returnNew) (michael) (grandma) (grandmaReturn) (jamesHome)] (travel) {};
    \node[font=\small, above] at (travel.north) {Travel Coordination};
    
    \node[group, fit=(turkeyPrep) (turkeyOven) (sarahLeaves) (turkeyDone) (sidePrep) (sideStart) (sideDone)] (food) {};
    \node[font=\small, above] at (food.north) {Food Preparation};
\end{pgfonlayer}
\end{tikzpicture}
\caption{Reactive Planning for Thanksgiving Dinner After James' Flight Delay, by Claude 3.7. Red boxes highlight constraint violations, including travel time, fire safety, side-dish preparation, and dinner deadline.}
\label{fig:thanksgiving-reactive-workflow-p9}
\end{figure*}
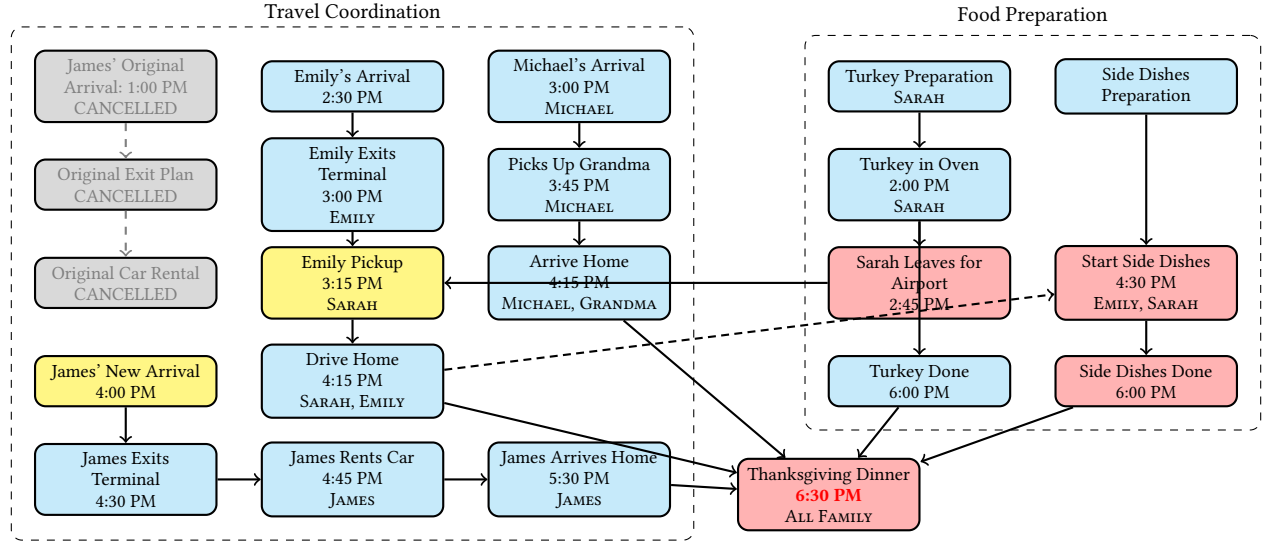

Problem~\textbf{P6} considers a Thanksgiving dinner scenario in which a family of five must return to their home in a Boston suburb for a 6~p.m.\ dinner. The problem involves coordinating departure times, managing travel logistics (including possible traffic delays), and ensuring timely arrival. Table~\ref{tab:ThanksgivingDinner1} formalizes these challenges as a sequential planning problem.
This scenario also lays the foundation for a more advanced disruption case, which has proven difficult for standalone LLMs, as discussed in~\textbf{P9}.

\subsubsection{Common Sense Augmentation}

Figure~\ref{fig:thanksgiving-workflow-p6} presents a feasible schedule planned by Claude 3.7. Similarly, GPT-4o was able to generate a viable plan to ensure dinner was started on time (figure is similar and therefore not shown). However, a subtle, yet important consideration that humans typically account for---but LLMs initially overlooked---is the time required for passengers to retrieve their luggage after landing. In practice, this process typically takes about 30 minutes before they exit the terminal.  

To address this, \emph{common-sense augmentation agent} was introduced into the plan. The yellow boxes in Figure~\ref{fig:thanksgiving-workflow-p6} reflect
this augmentation by introducing 30 minutes for James and Emily to exit the
airport.

\subsubsection{Context Narrowing}

Next, we use problem \textbf{P9} to illustrate the attention-narrowing problem and the importance of independent validation. Problem \textbf{P9} is identical to the previous instance, except that at 1 PM, James notifies the group that his plane will land at 4 PM instead of 1 PM due to an emergency detour.  
Figure~\ref{fig:thanksgiving-reactive-workflow-p9} shows that Claude 3.7's reactive planning introduces constraint violations:

\begin{itemize}[leftmargin=1em, topsep=-.05em, parsep=-.05em, label=-]
    \item \emph{Fire Safety}: Sarah is scheduled to leave home at 2:30 PM, leaving the oven unattended.
    \item \emph{Travel Time}: The travel time between home and BOS should be one hour, but is scheduled to be only 30 minutes.
    \item \emph{Side Dish Preparation}: The required preparation time is 2 hours, but only 90 minutes are allocated.
    \item \emph{Dinner Time}: Dinner is now scheduled for 6:30 PM, violating the 6:00 PM constraint.
\end{itemize}

Each of these violations is perplexing, given that the constraints are explicitly stated in the context. Furthermore, after multiple iterations of reactive planning within the same thread, several constraints continue to be ignored or misinterpreted (e.g., cooking safety). This highlights a key limitation in the model's ability to maintain global constraint awareness over sequential planning steps due to attention narrowing, presented in Section~\ref{sec:llm-limitations}.

When tested with GPT-o1, all constraints were correctly observed. However, in the final step, it added 30 additional minutes to James' driving time from Boston airport to home, citing potential traffic congestion. This kind of cleverness is, on the one hand, appreciated because the LLM injects common sense. However, it is also concerning, as an LLM may inject its own opinions at unpredictable stages in unpredictable ways. For common sense injection, human supervision would be preferable to ensure that the applied common sense reflects the shared understanding that is truly \emph{common} between people, particularly those living in the Boston area.

This pattern suggests that during reactive planning within the same thread, the model fixates on recent adjustments while progressively disregarding earlier constraints. According to \cite{liu2024largelanguagemodelsintrinsic}, some context may be lost randomly in the middle of the context buffer, further contributing to systematic attention narrowing and planning inconsistencies.


\subsubsection{\textbf{$\SAGA$ Remediation: Context Management \ and Reactive Planning}}

To address the issue of context narrowing and loss, $\SAGA$ employs global \emph{coordination and context management agents} (as depicted in Sections~\ref{sec:globalCoordination} and \ref{sec:contextmanagement}) to checkpoint historical state transitions, unresolved dependencies, and constraints. A key design criterion is to keep the agent's context small to prevent it from suffering from attention narrowing itself. Hence, we employ two validation agents: one for travel coordination and another for food preparation, each maintaining a context of less than $1k$.

The travel coordination agent records in external storage the temporal-spatial states of each individual and relevant temporal constraints. For problem \textbf{P9}, it stores the individual's \emph{current state}, \emph{next scheduled state-transition time}, and \emph{all relevant  constraints}.

When an unexpected event triggers reactive planning, all individuals roll back to the last saved state. The system then consolidates past and new constraints, resolving conflicts through ``compensational'' schedule cancellation before proceeding with rescheduling. This ensures that:

\begin{itemize}[leftmargin=1em, topsep=-.05em, parsep=-.05em, label=-]
\item \textit{Past history} is preserved and not inadvertently overridden.
\item \textit{New dependencies and constraints} are properly restored (e.g., oven safety watch) and integrated.
\item \textit{Consistency across state transitions} is maintained.
\end{itemize}

By maintaining a structured history of constraint awareness, $\SAGA$ ensures robust planning, effectively mitigating LLM-driven attention narrowing and enhancing consistency in reactive temporal scheduling.

\subsection{Wedding Gathering Problems: P5 and P8}
\label{sec:exp-p5p8}

Table~\ref{tab:Wedding} presents a wedding travel coordination problem (Problem \textbf{P5} in \cite{2025realmbench}). Several friends arrive at different times and locations before a 3:00~PM photo session. The challenge includes using two vehicles for airport pick-ups (for those unable to drive or saving costs) and completing key errands like collecting the wedding gift and retrieving formal attire. All activities must be scheduled to ensure timely arrival at the venue.

\begin{table}[t!]
\caption{Wedding Reunion Logistics Problem}
\centering
\begin{small}
\renewcommand{\arraystretch}{1.1}
\setlength{\fboxsep}{5pt}
\fbox{
\begin{minipage}{0.45\textwidth}
\textbf{Metrics:}
\begin{itemize}[leftmargin=1em, topsep=-.1pt, itemsep=-.1pt, label=-]
\item \textbf{On-time performance:} Must arrive at the venue for 3:00 PM photos.
\end{itemize}
\textbf{Locations:} Four locations: $V = \{B, G, T, W\}$, where $B$ is Boston Airport, $G$ is Gift shop, $T$ is Tailor shop, and $W$ is Wedding venue.

\textbf{Travel time:} (minutes) 

$~B\text{-}G:45, ~B\text{-}T:30, ~B\text{-}W:40, ~G\text{-}T:20, ~G\text{-}W:25, ~T\text{-}W:15$. 

\textbf{Arrival Times:}  
\begin{itemize}[leftmargin=1em, topsep=-.1pt, itemsep=-.1pt, label=-]
\item Alex: At $B$ at 11:00 AM from Chicago (need a ride)
\item Jamie: At $B$ at 12:30 PM from Atlanta (need a ride)
\item Pat: At $W$ at 12:00 PM driving from NYC (has 5-seater car)
\end{itemize}

\textbf{Required Tasks:}
\begin{itemize}[leftmargin=1em, topsep=-.1pt, itemsep=-.1pt, label=-]
    \item Gift collection from $G$ (after 12:00 PM)
    \item Clothes pickup from $T$ (by 2:00 PM)
    \item Photos at $W$ (3:00 PM sharp)
\end{itemize}

\textbf{Available Resources:} 
\begin{itemize}[leftmargin=1em, topsep=-.1pt, itemsep=-.1pt, label=-]
    \item One car (5-seater) with Pat, available after he is Boston
    \item Local friend Chris (5-seater) available after 1:30 PM at $W$
\end{itemize}

\textbf{Scheduling Constraints:}  
- All tasks must complete before 3:00 PM photo time
- Gift store opens at 12:00 PM
- Tailor closes at 2:00 PM
- Two cars must accommodate all transport needs
\end{minipage}
}
\label{tab:Wedding}
\end{small}
\vspace{-.18in}
\end{table}

\begin{table}[th!]
\centering
\caption{Wedding Reunion Logistics Schedule, by Claude 3.7. (Planning error rows in red)}
\begin{small}
\begin{tabular}{>{\raggedright\arraybackslash}p{1.36cm} >{\raggedright\arraybackslash}p{4.6cm} >{\raggedright\arraybackslash}p{1.3cm}}
\toprule
\textbf{Time} & \textbf{Activity} & \textbf{People} \\
\midrule
\rowcolor{gray!10} 11:00 AM & Alex arrives at Boston Airport (B) & Alex \\
12:00 PM & Pat arrives at Wedding Venue (W) & Pat \\
\rowcolor{gray!10} 12:00 PM & Gift Shop (G) opens & -- \\
12:00--12:40 & Pat drives from Wedding Venue (W) to Boston Airport (B) & Pat \\
\rowcolor{gray!10} 12:30 & Jamie arrives at Boston Airport (B) & Jamie \\
12:40--12:45 & Pat picks up Alex at Boston Airport (B) & Pat, Alex \\
\rowcolor{gray!10} 12:45--12:50 & Pat picks up Jamie at Boston Airport (B) & Pat, A., J.  \\
12:50--1:35 & Drive from BOS to Gift Shop (G) & Pat, A., J. \\
\rowcolor{gray!10} 1:30 & Chris available at Wedding Venue (W) & Chris \\
1:35--1:50 & Collect gift at Gift Shop (G) & Pat, A., J.  \\
\rowcolor{gray!10} 1:50--2:10 & Drive from Gift Shop (G) to (T) & Pat, A., J.  \\
\rowcolor{red!20} 2:00  & Tailor Shop (T) closes & -- \\
\rowcolor{red!20} 2:10--2:25  & Pick up clothes at Tailor Shop (T) & Pat, A., J.  \\
2:25--2:40 & Drive from T to Wedding Venue (W) & Pat, A., J.  \\
\rowcolor{gray!10} 2:40 PM & Arrive at Wedding Venue (W) & Pat, A., J.  \\
3:00 PM & \cellcolor{pink!15}Photo session at Wedding Venue (W) & All \\
\bottomrule
\end{tabular}
\end{small}
\label{tab:wedding-claude}
\end{table}

\subsubsection{Context Narrowing (again)}

Table~\ref{tab:wedding-claude} presents an infeasible schedule generated by Claude 3.7, where Pat arrives at the tailor shop (T) after closing time: another example of attention narrowing. When queried about the error, Claude 3.7 admitted that it prioritized local route optimization while losing track of global constraints.  

To remedy this issue, $\SAGA$ can enforce constraint validation checkpoints at 12:50 PM (evaluating whether to send Pat to T) or at 1:30 PM (when Chris becomes available to drive to T). These missed optimization opportunities can be addressed through $\SAGA$'s validation protocols.

In contrast, GPT-o1 correctly schedules Pat to visit the tailor shop (T) first, ensuring it is open, before proceeding to the gift shop (G) and successfully completing both errands. (Due to space limitations, we do not present the successful results.)

However, both schedules overlook a more efficient alternative: Chris, who is available at 1:30 PM, could have handled both errands, balancing the workload and improving overall efficiency. The comparative travel routes for Pat and Chris are:

\begin{itemize}[leftmargin=1em, topsep=0pt, itemsep=0pt, label=-]
    \item Pat's route: W→B (40 min) + B→W (40 min) = 80 minutes.
    \item Chris's route: W→T (15 min) + T→G (20 min) + G→W (25 min) = 60 minutes.
\end{itemize}

\subsubsection{Transaction Properties}

Problem \textbf{P8} introduces a traffic alert:

\noindent
{\color{black}{\textbf{Alert 1:00 PM:}}} Traffic Alert, an accident near Logan Airport in Boston \textbf{triples} all travel times to and from the airport!
\textbf{Only 
$\SAGA$ correctly handles this alert.}

\begin{table}[th!]
\centering
\caption{Claude 3.7 Ignored Traffic Delay (errors in red)}
\vspace{-.1in}
\begin{small}
\begin{tabular}{>{\raggedright\arraybackslash}p{1.36cm} >{\raggedright\arraybackslash}p{4.6cm} >{\raggedright\arraybackslash}p{1.3cm}}
\toprule
\textbf{Time} & \textbf{Activity} & \textbf{People} \\
\midrule
\rowcolor{gray!10} 1:00 & {Traffic Alert: Accident near Airport triples travel times to/from airport} & -- \\
1:00 & Current status: Pat, Alex, and Jamie en route from Airport (B) to Tailor (T) & Pat, A., J. \\
\rowcolor{red!20} 1:00--1:10 & Emergency decision: Continue to  (T) & Pat, A., J. \\
\rowcolor{red!20} 1:10--1:25 & Arrive at (T), collect clothes & Pat, A., J. \\
1:25--1:45 & Travel from Tailor (T) to Gift Shop (G) & Pat, A., J. \\
1:30 & Chris is available at (W) & Chris \\
\rowcolor{gray!10} 1:30--1:45 & Chris drives from (W) to (G) & Chris \\
1:45--2:00 & Both cars meet at (G), collect gift & P., A., J., C \\
\rowcolor{gray!10} 2:00--2:25 & Pat's car: Drive from (G) to (W) & P., A., J. \\
2:00--2:25 & Chris' car: Drive from (G) to (W) & Chris \\
\rowcolor{gray!10} 2:25 & All arrive at Wedding Venue (W) & P., A., J., C. \\
3:00 & \cellcolor{white!15}Photo session at Wedding Venue (W) & All \\
\bottomrule
\end{tabular}
\end{small}
\label{fig:reactive-Claude}
\vspace{-.1in}
\end{table}

\begin{table}[thbp]
\centering
\caption{DeepSeek's Failed Reactive Schedule After Traffic Alert, and GPT-4o Made Similar Errors (errors in red)}
\begin{small}
\begin{tabular}{>{\raggedright\arraybackslash}p{1.36cm} >{\raggedright\arraybackslash}p{4.6cm} >{\raggedright\arraybackslash}p{1.3cm}}
\toprule
\textbf{Time} & \textbf{Activity} & \textbf{People} \\
\midrule
\rowcolor{red!20} 1:00 & Traffic alert received - Pat at W & System \\
\rowcolor{red!20} 1:05 & Pat departs W for B & Pat \\
1:30 & Chris becomes available at W & Chris \\
\rowcolor{gray!10} 1:30 & Chris departs W for T & Chris \\
1:45 & Chris arrives at T for clothes & Chris \\
\rowcolor{gray!10} 2:00 & Chris departs T with clothes & Chris \\
2:15 & Chris arrives at G for gifts & Chris \\
\rowcolor{gray!10} 2:25 & Pat arrives at B (delayed by traffic) & Pat \\
2:35 & Pat departs B with Alex \& Jamie & Pat \\
\rowcolor{gray!10} 2:40 & Chris departs G with gifts & Chris \\
2:55 & Chris arrives at W & Chris \\
\rowcolor{red!20} 3:55 & Pat's group arrives at W (Late) & Pat \\
\bottomrule
\end{tabular}
\end{small}
\label{fig:failed-reactive-DS}
\end{table}

This alert requires LLMs to replan in real time. Unfortunately, 
{Claude~3.7}, {DeepSeek~R1}, and {GPT-4o} failed 
to react accurately to the new traffic constraints, and even {GPT-o1} struggled with the precision of the planning. In contrast, 
$\SAGA$ can help remedy these shortcomings by maintaining both
transaction state and history.

The following is a list of results from four LLMs:

\begin{itemize}[leftmargin=1.2em, topsep=.05em, parsep=.05em, label=*]
\item \textbf{Claude}:
Table~\ref{fig:reactive-Claude} shows that Claude 3.7 recognizes the accident, but does not update Pat's driving time from Boston Airport (departing at 12:50\,PM) to the gift shop. In other words, Claude 3.7 fails to fully transition into the new alert state.

\item \textbf{DeepSeek R1}:
Table~\ref{fig:failed-reactive-DS} demonstrates how DeepSeek~R1 fails to maintain temporal consistency in reactive planning. When the traffic alert takes effect at 1:00\,PM, DeepSeek discards its execution history and attempts to create a new plan starting from that point onward. Critically, it reassigns Pat to begin driving to the airport at 1:00\,PM, even though Pat had already arrived at the airport by 12:40\,PM under the original schedule. This ``rewrite'' of already-executed actions illustrates how LLMs can lose track of immutable past events when adapting to new conditions.

\item \textbf{GPT-4o}:
Similar to DeepSeek~R1, GPT-4o exhibits temporal-spatial 
context confusion and violates multiple constraints, 
demonstrating that it struggles to adapt effectively once alerts 
are introduced mid-plan. 
\textit{(Table not shown due to space limitations.)}

\item \textbf{GPT-o1}:
Table~\ref{tab:wedding-GPT-o1-good} shows GPT-o1's \emph{conservative} plan in which Chris handles the tailor shop, avoiding potential delays for Pat. The solution is feasible but coarse-grained, as it doesn't leverage precise spatial-temporal reasoning about Pat's current position relative to the accident location. A more refined approach would first determine whether Pat has already passed the accident site by 1:00~PM, which could eliminate unnecessary detours and resource reallocations. This highlights the difference between merely finding a feasible solution versus optimizing based on detailed state information.
\end{itemize}

\begin{table}[th!]
\centering
\caption{Wedding Reunion Reactive Schedule, by GPT-o1}
\vspace{-.1in}
\begin{small}
\begin{tabular}{%
    >{\raggedright\arraybackslash}p{1.36cm}
    >{\raggedright\arraybackslash}p{4.6cm}
    >{\raggedright\arraybackslash}p{1.3cm}}
\toprule
\textbf{Time} & \textbf{Activity} & \textbf{People} \\
\midrule
\rowcolor{gray!10}
11:00 AM 
  & Alex arrives at Airport (B). 
  & Alex \\
12:00 PM 
  & Pat departs for Airport from (W). 
  & Pat \\
\rowcolor{gray!10}
12:30 PM 
  & Jamie arrives at Airport (B). 
  & Jamie \\
12:40--12:50 PM 
  & Pat arrives at (B), picks up Alex and Jamie; 
    departs at 12:50 PM. 
  & Pat, A., J. \\
\rowcolor{blue!10}
\textbf{12:50--1:00 PM} 
  & \textbf{Drive (B $\to$ W) under normal conditions 
    for first 10 minutes.} 
  & Pat, A., J.  \\
\rowcolor{red!20}
\textbf{1:00--2:30 PM} 
  & \textbf{Traffic Alert starts: 
    remaining distance (30 min normal) 
    becomes 90 min. 
    Arrival at W by 2:30 PM.}
  & Pat, Alex, Jamie \\
\rowcolor{gray!10}
1:30 PM 
  & Chris available at (W). Departs for Tailor. 
  & Chris \\
1:30--1:45  
  & Drive (W $\to$ T). 
  & Chris \\
\rowcolor{gray!10}
1:45--1:50 
  & Pick up clothes at (T), closes at 2:00 PM. 
  & Chris \\
1:50--2:10  
  & Drive (T $\to$ G). 
  & Chris \\
\rowcolor{gray!10}
2:10--2:15 
  & Purchase gift at (G). 
  & Chris \\
2:15--2:40  
  & Drive (G $\to$ W). 
  & Chris \\
\rowcolor{gray!10}
2:30 PM 
  & Pat, Alex, Jamie arrive at (W). 
  & Pat, A., J.  \\
2:40 PM 
  & Chris back at (W) with clothes and gift. 
  & Chris \\
\textbf{3:00 PM} 
  & Wedding photo session at (W). 
  & Everyone \\
\bottomrule
\end{tabular}
\end{small}
\label{tab:wedding-GPT-o1-good}
\end{table}

\subsubsection{\textbf{LLM Limitations and $\SAGA$ Remediation}}

This study reveals critical limitations in how modern LLMs handle disruptions in planning scenarios:

\begin{itemize}[leftmargin=1.0em, topsep=.05em, parsep=.05em, label=-]
   \item \textbf{State Maintenance Failure:} 
   When an alert occurs, these LLMs might discard the partial context of already-completed actions, attempting to generate entirely new plans rather than adapt existing ones. This reveals their inability to reason about the continuous flow of time in real-world scenarios.
   \item \textbf{Temporal Inconsistency:} They attempt to modify immutable past events.
   \item \textbf{Position Tracking:} Agent locations are lost at critical intervals.
   \item \textbf{Path Dependency:} Models cannot recognize that different segments of a journey may be differently impacted by an alert.
\end{itemize}

By contrast, $\SAGA$ implements a comprehensive remediation approach through fine-grained compensation:

\begin{itemize}[leftmargin=1.0em, topsep=.05em, parsep=.05em, label=-]
    \item \textbf{Persistent Context Repository:} $\SAGA$ maintains an external state repository that captures the complete world state at each checkpoint, enabling reliable rollback and forward projection regardless of attention constraints in the planning agent.
    
    \item \textbf{Immutable Action Logging:} All executed actions are recorded as immutable transactions in a persistent log, ensuring that historical events remain consistent even when replanning occurs, preventing the ``amnesia effect'' common in LLM planners.
    
    \item \textbf{Compensatory Planning:} When disruptions occur, $\SAGA$ doesn't simply replan from scratch but applies compensatory actions specifically designed to address the deviation while preserving as much of the original plan as possible.
    
    \item \textbf{Constraint Consistency Validation:} The system continuously validates that new plans remain consistent with both physical limitations and temporal dependencies established in earlier planning phases.
\end{itemize}

\paragraph{$\SAGA$ Compensatory Analysis:} When faced with disruptions (e.g., the 1:00 PM traffic alert in our wedding scenario), $\SAGA$ executes a structured compensation process:

\vspace{-.1in}
\begin{align}
T_{\text{affected}} &= \max(0, T_{\text{total}} - T_{\text{elapsed}})\\
T_{\text{new}} &= T_{\text{elapsed}} + (M \cdot T_{\text{affected}})
\end{align}
\vspace{-.1in}

This approach enables three key capabilities: (1) partial journey compensation for route segments, (2) strategic resource reallocation when needed, and (3) principled constraint relaxation with appropriate compensatory actions. Here, the key state to facilitate a precise resolution is to answer the question: ``Has Pat's vehicle passed the accident location at 1:00 PM (and hence unaffected)?'' The answer determines the remaining time required to reach the originally scheduled destination, the Tailor. In such a case, no rescheduling is required. If Pat's car is unfortunately involved in the accident, a more comprehensive replanning approach would be necessary to accommodate this significant disruption.

\subsection{Observations}

Our experiments across multiple LLMs (GPT-o1, DeepSeek~R1, Claude~3.7, GPT-4o) highlight consistent limitations in complex planning scenarios. While GPT-o1 showed partial historical awareness, all models exhibited attention narrowing, self-validation failure, and inconsistent spatial-temporal reasoning.

Table~\ref{tab:LLMvsSAGA} summarizes $\SAGA$'s context management and compensation mechanisms directly address these limitations of LLMs.

\begin{table}[t!]
\centering
\small
\caption{LLMs vs. $\SAGA$ on Context Management}
\vspace{-.1in}
\begin{tabular}{lcc}
\toprule
\textbf{Capability} & \textbf{Standard LLMs} & \textbf{$\SAGA$} \\
\hline
\midrule
Maintains historical actions & Partial/None & Full \\
Partial journey compensation & Rarely & Always \\
Constraint consistency checking & Ad-hoc & Systematic \\
Handles attention narrowing & Vulnerable & Resistant \\
Physical-temporal consistency & Inconsistent & Guaranteed \\
\bottomrule
\end{tabular}
\label{tab:LLMvsSAGA}
\end{table}

\section{Conclusion}
\label{sec:conc}

We introduced $\SAGA$, a structured transactional multi-agent framework that addresses four fundamental limitations of existing LLM-based planning systems: inadequate self-validation, context narrowing, absence of transaction properties, and insufficient inter-agent coordination.

Our experiments demonstrate that even advanced LLMs like Claude 3.7 and GPT-o1 often fixate on recent context while neglecting critical earlier constraints, particularly in reactive planning scenarios where models attempt to retroactively rewrite past actions rather than adapting from the current state.

$\SAGA$ overcomes these limitations with four key innovations:
\begin{enumerate}[leftmargin=1.36em, topsep=.15em, parsep=.15em, label=\arabic*.]
\item \textbf{Independent validation} to address self-validation gaps
\item \textbf{Strategic context preservation} to mitigate context narrowing
\item \textbf{Transactional state management} with immutable records and compensation mechanisms
\item \textbf{Specialized agent coordination} with explicit role distribution and dependency tracking
\end{enumerate}

These innovations enable robust planning across diverse scenarios, from travel logistics to dynamic tasks. By enforcing rigorous transactional validation among specialized agents, $\SAGA$ significantly improves consistency, reliability, and adaptability for mission-critical applications. Due to space constraints in this experience paper, detailed algorithm specifications are provided in our companion ALAS paper \cite{chang2025ALAS}.

Future work will address several critical challenges: developing formal verification methods for LLM-generated compensation code to ensure logging and rollback correctness, addressing intrinsic autoregressive context limitations, creating comprehensive verification frameworks for transactional multi-agent systems, and extending $\SAGA$ to domains requiring scientific reasoning \cite{SocreticMethodChang2023, AGIBookChang2024, UCCTchang2025}, creative collaboration, and decision-making under uncertainty. 

\section*{Acknowledgment}

I am deeply grateful to my late advisor, Hector Garcia-Molina, for his invaluable mentorship and for pioneering SAGAS, the foundational inspiration for this work.

\newpage
\balance
\bibliographystyle{plainnat}
\bibliography{TemporalPlanning, EdwardChang, workflow, TSP}

\end{document}